%% file: main.tex
\documentclass[runningheads]{llncs}

\PassOptionsToPackage{table}{xcolor}  % must precede eccv.sty, which loads xcolor
\usepackage{eccv2026/eccv}

\usepackage{eccv2026/eccvabbrv}

\usepackage{graphicx}
\usepackage{booktabs}

\usepackage[accsupp]{axessibility}  % Improves PDF readability for those with disabilities.

\input{preamble}

\usepackage{hyperref}

\usepackage{orcidlink}

\begin{document}

% ---------------------------------------------------------------
\title{Anchored, Not Graded: Vision-Language Models\\Fail at Slant-from-Texture Perception}

% Abbreviated title for running head
\titlerunning{VLMs Fail at Slant-from-Texture Perception}

% TODO FINAL: Add \orcidlink{} entries for camera-ready version
\author{Qian Zhang\inst{1} \and
Michal Golovanevsky\inst{1,3}\and
Fulvio Domini\inst{2}\and
James Tompkin\inst{1}}

\institute{Brown University Computer Science \and Brown University Cognitive Science \and Harvard University}
% \institute{Brown University, % \institute{Brown University, Providence RI 02912, USA \and Harvard University, Boston MA 02135, USA}

\authorrunning{Q.~Zhang et al.}

\maketitle

\setcounter{footnote}{0}

\input{0_abstract}
\input{1_intro}
\input{3_method}
\input{4_results}
\input{4_results2_vit}
\input{4_results3_lm}
\input{5_conclusion}
\input{2_related}

\paragraph{Acknowledgements:}

QZ, JT thank NSF CAREER 2144956, NSF IIS-2107409, and an Andy van Dam Fellowship. We are grateful to the anonymous reviewers for their thoughtful comments, and we thank Zitian Tang, Zhenyu Zhu, Channy Lim, Mikhail Okunev, Daniel Ritchie, and Thomas Serre for insightful discussions.

{
\small
\bibliographystyle{splncs04}
\bibliography{main}
}

% WARNING: submit supplementary separately from main paper
\clearpage
\appendix
\input{7_supplemental}
% \input{6_todos}

\end{document}

%% file: preamble.tex
\usepackage[dvipsnames]{xcolor}
\newcommand{\TODO}[1]{\textbf{\color{magenta}[TODO: #1]}}
\newcommand{\CC}[1]{\textbf{\color{magenta}[Edit:} #1 \textbf{\color{magenta}]}}
\newcommand{\QZ}[1]{\textcolor{magenta}{{[Qian: #1]}}}
\newcommand{\JT}[1]{\textcolor{red}{{[James: #1]}}}
\newcommand{\MG}[1]{\textcolor{violet}{{[Michal: #1]}}}

\renewcommand{\TODO}[1]{}
\renewcommand{\QZ}[1]{}
\renewcommand{\JT}[1]{}
\renewcommand{\MG}[1]{}
\renewcommand{\CC}[1]{#1}

\usepackage[table]{xcolor}

\usepackage{array}

\definecolor{gray}{gray}{0.7}
\definecolor{codegreen}{RGB}{65,133,244} % similar to ForestGreen

\newcolumntype{C}[1]{>{\color{codegreen}\ttfamily}p{#1}}

\usepackage{longtable}

\usepackage{siunitx}

\usepackage{textcomp, gensymb} % provides command \degree

\usepackage{multirow}

\usepackage{microtype}
\usepackage[subtle]{savetrees}

\usepackage{enumitem}

%% file: 0_abstract.tex
\begin{abstract}
Human perception of surface slant from texture exhibits systematic, graded biases that emerge reliably in psychophysical experiments. Prior work showed that unsupervised CNNs reproduce several human-like biases, while supervised CNNs do not. Do Vision-Language Models (VLMs) exhibit similar competencies? Across multiple VLM families and model scales, zero-shot and in-context prompting both produce distinctive failures: slant is predicted at only a small set of anchors (e.g., 0\degree, $\pm$45\degree) with little dependence on stimulus slant, field of view, or surface curvature. Supervised fine-tuning partially remediates the failure, but residual anchoring persists.
%predictions become monotonically related to ground truth and curvature-sign accuracy improves, but residual anchoring persists. %Our results establish that VLMs differ qualitatively from humans and unsupervised CNNs: where humans exhibit smooth, stimulus-dependent biases, VLMs produce discrete, stimulus-insensitive outputs. 
While success in high-level vision-language benchmarks might not require sensitivity to low-level geometric cues, we interpret anchoring as a failure at the representation-to-output language interface: not necessarily an absence of geometric encoding, but a failure to express it in a graded form. 
% Code and dataset will be released soon.
\end{abstract}
\vspace{-0.5cm}

% TODO: make a foot note here:
% code gitrepo:
% dataset HF:

%% file: 1_intro.tex
\vspace{-0.25cm}
\section{Introduction}
\vspace{-0.15cm}

Comparing human and neural network visual systems has a long anecdotal history, but recent large models trained on billions of images have made the comparison of their behaviors and mechanisms more meaningful. Understanding these relations through \emph{in silico} experiments is an important direction for vision science, e.g., to raise hypotheses for how biological vision might function. But it is also useful for practical tasks. Often, human-AI collaboration is predicated on AI systems being able to predict what humans perceive (AI: \emph{``I can see that.''} Human: \emph{``I cannot see that.''}). Yet, studies of whether the basic psychophysical abilities of AI systems match those of humans are rarer, as most works focus on downstream performance.

Take the simple task of estimating the slant of a surface. Humans are able to judge three-dimensional surface slant from texture gradients alone, yet these judgments are systematically biased rather than accurate. Psychophysical experiments using textured slant surfaces have identified at least three consistent biases in perception~\cite{todd2005effects}: (B1) convex surfaces appear steeper than concave surfaces of equal physical slant; (B2) larger fields of view produce greater perceived slant; (B3) curvature-sign discrimination degrades at small fields of view, falling to chance at 5\degree~FOV.
% ; and (B4) regular textures elicit stronger slant impressions than irregular ones. 
%\MG{I would rephrase "veridical" to "accurate" or "grounded", since the rest of the paragraph is terminology-heavy. Also sentence should say "four" or "three" biases?}
%\JT{The word 'veridical' comes from copsy language; we can go with 'accurate'. 'Grounded' suggests language, which is not the right meaning.}
%
These biases are commonly quantified by perceptual gain: the ratio of judged slant to ground-truth slant. This is approximately 0.56 for regular dot textures, indicating substantial underestimation of surface slant. Curvature-sign accuracy for such textures averages around 86\% overall but drops sharply at narrow fields of view. Thus, human slant-from-texture perception reflects systematic image-level heuristics, such as sensitivity to texture scaling gradients, rather than recovery of precise surface geometry.

Do neural networks exhibit similar biases? Wang et al.~\cite{wang2023human} investigated this question using convolutional neural networks (CNNs) trained on synthetic dot-textured stimuli. Unsupervised CNN autoencoders trained with a reconstruction objective reproduced human-like slant-from-texture biases. Human-like error patterns were recovered from the CNN's internal representations using simple linear projections, which links representational structure to perceptual behavior. However, CNNs trained with direct supervision to regress physical slant achieved unbiased performance. This dissociation indicates that human-like biases arise not from architecture alone, but from learning objectives that emphasize texture statistics rather than explicit geometric labels. These results raise the question of whether such biases persist in visual representations both as models move beyond convolutional architectures and as they are trained under fundamentally different objectives.

\input{figures/teaser.tex}

Vision-Language Models (VLMs) present such a case. These models combine large-scale vision encoders, such as Vision Transformers, pretrained on broad image distributions, with language models trained on large-scale text corpora. Each is aligned through multimodal supervision on image-caption pairs. The architectural differences from CNNs and the distinct training routine are often assumed to yield representations that are richer or more semantically structured \cite{dosovitskiy2020image, zhang2024vision, zhai2021lit, radford2021learning}. 
Recent work has begun to question the geometric ability of VLMs in other domains~\cite{rudman2025forgotten, huang2025vision}, but no studies have examined texture-based slant perception. 
% or compared VLM behavior to human psychophysical baselines.\JT{This last claim is not true afaik.}
Such stimuli contain no objects or semantic content; they contain only texture gradients that must be interpreted geometrically. Accordingly, despite strong performance on a range of vision-language benchmarks \cite{li2024llava, kamath2025gemma, bai2023qwen}, it remains unknown whether VLMs exploit low-level texture cues in a manner comparable to human perception or unsupervised vision models and whether they can communicate slant in language.

We evaluate multiple VLM families on a classical slant-from-texture task, using stimuli and ground truth matched to prior human and CNN studies.
We ask two questions. First, behavioral: do VLMs exhibit the same systematic biases as humans and unsupervised CNNs, e.g., convex-concave asymmetry (B1) and field-of-view effects (B2, B3). 
We find that VLMs in zero-shot settings exhibit strong \emph{anchoring} to a small set of discrete values, with predictions failing to exhibit monotonic dependence on stimulus parameters such as optical slant, field of view, or curvature sign---unlike the systematic biases observed in humans and unsupervised CNNs.
Second, intervention: can standard adaptation techniques---prompt engineering, in-context learning with labeled examples, or supervised fine-tuning---induce such biases? 
We show that prompt engineering and in-context learning do not help. Supervised fine-tuning does introduce a monotonic relationship between predicted and ground-truth slant and improves curvature-sign discrimination, but does not eliminate anchoring: predictions remain clustered at discrete values rather than varying continuously.
Probing the vision module within four VLMs with different architectures reveals a strong correlation between features and geometric quantities, suggesting that response anchoring is a language ``readout'' problem.
Together, these findings establish slant-from-texture as a \emph{diagnostic} task for revealing systematic differences in how humans, unsupervised vision models, and language-supervised multimodal models respond to texture-based geometric cues, and show where effort is needed to produce future models that can predict (as needed) the human response to the world around us.

\vspace{-0.1cm}
\paragraph{Assumptions and limitations.} 1) Our experiments cover slant from random polka dot textures only, rather than a broader space of textures that includes regularity and shape variation~\cite{todd2010perception} or natural image textures. 
As psychophysical studies have shown a strong response in humans and CNNs to our polka dot slant stimuli, 
% they are sufficient to establish VLM limitations~\QZ{From R1: "I would disagree with the statement that they are 'sufficient to establish VLM limitations' unless stronger evidence for this claim is demonstrated."}. 
they are useful probes to reveal existing VLM limitations.
2) We evaluate a breadth of models to characterize the landscape of VLMs, but specific alternatives may vary in significant ways. 3) Our findings do not obviate the usefulness of VLMs in end tasks or imply that VLMs must operate like humans; instead, they show differences in current VLM predictions with respect to human judgment that should be known when applying VLMs to tasks that require similar judgments.

% First, we do not manipulate texture regularity. Prior work shows that regular textures elicit stronger slant impressions than irregular ones (B4), and extending the present analysis to varying levels of texture regularity (jittering of dot locations) remains an important direction. Also, our conclusions do not yet extend to variations in element shape or contour regularity examined in prior psychophysical studies.

% Second, our behavioral comparison is limited to slant magnitude and curvature sign; other aspects of human shape-from-texture perception, such as perceived depth profiles across curved surfaces or judgments of curvature magnitude, especially when the homogeneous texture assumption is violated, are not examined~\cite{todd2010perception}.

%% file: figures/teaser.tex
\begin{figure}[t]
\centering
    \includegraphics[width=1.0\linewidth]{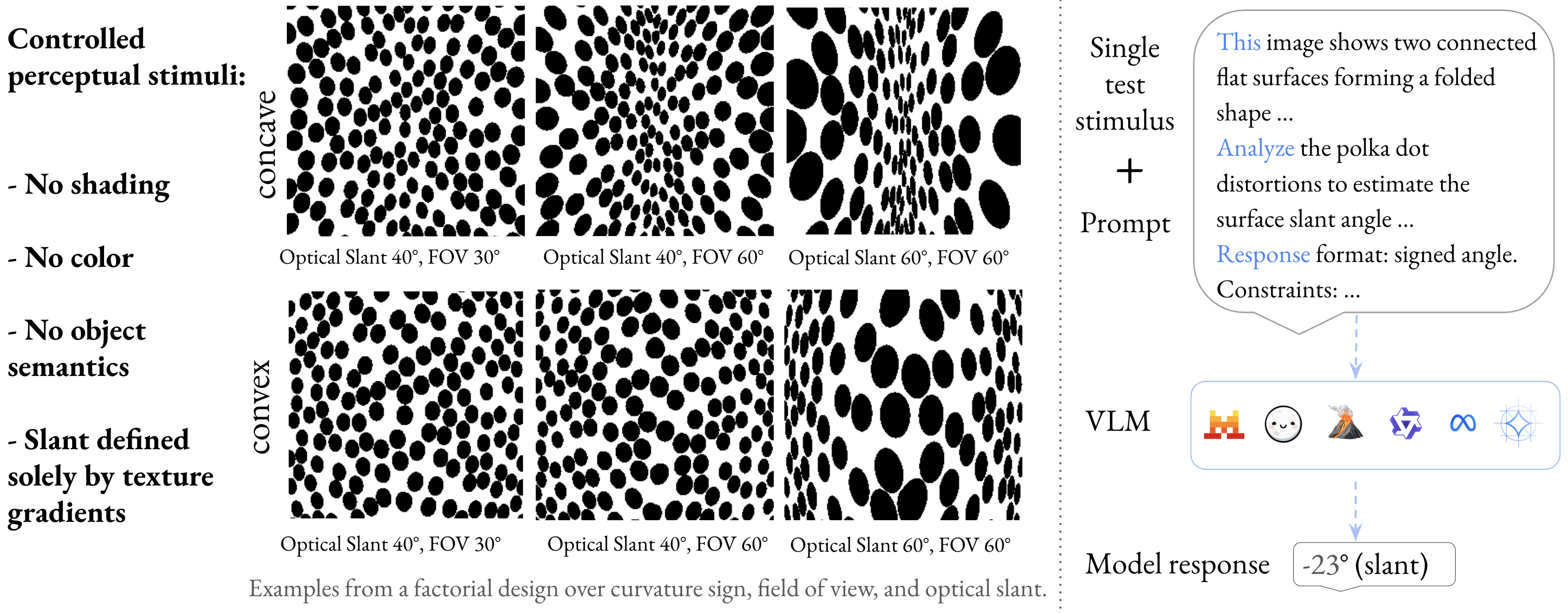}
\vspace{-0.5cm}
\caption{\textbf{Slant-from-texture as a controlled perceptual task for VLMs.}
\emph{Left:} Synthetic dot-textured surfaces vary in optical slant, field of view, and curvature sign. Texture gradients are the only cue to 3D orientation.
\emph{Right:} Given a single image and instruction, we ask VLMs to estimate surface slant and curvature sign from texture alone.
}
\label{fig:teaser}
\vspace{-0.7cm}
\end{figure}

%% file: 3_method.tex
\vspace{-0.25cm}
\section{Dataset and Experiment Setup}
\vspace{-0.15cm}

\paragraph{Stimuli.}
We produce synthetic dot-textured surfaces following Todd et al.~\cite{todd2005effects,todd2007effects}, and as followed by Wang et al.~\cite{wang2023human} (\cref{fig:teaser}). These have no shading, color, or silhouette cues---slant must be perceived from texture gradients. 
% See \cref{sec:related} for the dot texture-choice rationale.
% Todd et al.\ 2005 did experiments with different kinds of texture patterns, including plaid, regular contour, irregular contour, regular blob (polka dot), and irregular blob. The figures in their paper show a combined performance. Text in paper sometimes provides breakdown by texture type, but not always.
%
Each stimulus depicts two textured flat surfaces forming a dihedral angle, varying along four parameters:
\begin{center}
\begin{tabular}{l l}
\toprule
Optical slant $\sigma_{\text{cen}}$ & 10 values from 25\degree~to 60\degree \\
Field of view (FOV) & 10 values from 5\degree~to 60\degree \\
Curvature sign & Convex or concave \\
Texture & 12 i.i.d.~random dot jitters (with different random seeds) \\
\bottomrule
\end{tabular}
\end{center}

The surfaces have a true physical slant $\rho$ that is their angle relative to the fronto-parallel plane. It is derived from $\sigma_{cen}$ and FOV to preserve perceptual uniformity:
\[
\rho = \sigma_{\text{cen}} - s \cdot \frac{\text{FOV}}{4}, \hspace{0.5cm} s = -1\ \text{(concave)}, \hspace{0.5cm} s = +1\ \text{(convex)}.
\]

% \[
% \rho =
% \begin{cases}
% \sigma_{\text{cen}} + \tfrac{\text{FOV}}{4}, & \text{concave}, \\[6pt]
% \sigma_{\text{cen}} - \tfrac{\text{FOV}}{4}, & \text{convex}.
% \end{cases}
% \]
Optical slant $\sigma_{\text{cen}}$ is the controlled experimental parameter matched across curvature conditions; physical slant $\rho$ is what observers judge. Optical slant corresponds directly to the texture gradients while physical slant requires integrating curvature and FOV information. This means physical slant prediction require geometric understanding of 3D structure rather than a direct ``readout'' of local texture cues.
% \QZ{But we do not give FOV as input info in prompt, whereas human can infer this by changing of viewing distance.}

The first three attributes form a 200-condition factorial design (10 optical slant \texttimes 10 FOV \texttimes 2 curvature sign). 
%Textures were randomly generated for each image i.i.d., ensuring no two images were identical.
As each instantiation of the 200 conditions has a random texture, each stimulus provides a new projection of the same geometry. Geometric reasoning requires an understanding of distortions to slant generalizable over appearance noise from texture jitter.
We split the stimuli into two sets: 2000 for any training (fine-tuning, in-context learning, etc.) and 400 for testing.
% \QZ{Document the split strategy: random or stratified by condition? A reviewer will ask.}

\paragraph{Prompts.}
Each stimulus is paired with a text prompt. We vary the information provided in the prompt and its style to try to avoid any particular pitfall in how we describe the problem in language. This design allows us to test whether, and how, linguistic framing modulates visual estimations. The prompt varies the task, whether additional cues are given (e.g., \emph{``The greater the slant angle, the more distorted the dots will appear.''}), and the style of prompt in its input and output. This produces seven variants for physical slant angle prediction and three for binary concave/convex prediction (\cref{tab:mainpaper_prompt_variants_natural}). We state all prompts in the supplementary material.
% \cref{tab:prompt_natural}, \cref{tab:prompt_technical}, and \cref{tab:prompt_binary}.
\begin{center}
\vspace{-0.4cm}
\resizebox{\linewidth}{!}{
\begin{tabular}{l l}
\toprule
Task type & Slant prediction (regression) vs.\ curvature sign (binary classification). \\
Instruction detail & Minimal vs.\ enriched with cue descriptions. \\
\midrule
Language style & Natural (colloquial, e.g., ``ridge/valley'', ``fold'') vs. \\ 
& technical (e.g., ``concave/convex'', ``dihedral angle''). \\
Output format & Free text vs.\ structured JSON. \\
\midrule
In-context & Absence vs.\ presence of example stimuli with slant labels. \\
% \item Additional information: explicit references to known human/CNN biases, or supervised fine-tuning with labeled data.
\bottomrule
\end{tabular}
}
\end{center}

Additionally, we provide an in-context learning prompt modifier. Along with the test stimuli, such prompts include four labeled example stimuli from the training set that are balanced across slant angle, FOV, and curvature sign. Labels are provided as free text, details in the supplementary material.
% \cref{tab:prompt_incontext}).

\input{tables/prompt_natural_binary.tex}

\paragraph{Model Families and Runs.}
We evaluated six recent open-source VLMs (Gemma3~\cite{kamath2025gemma}, LLaMA4~\cite{llama4_ollama}, LLaVA1.6~\cite{li2024llava}, Mistral~\cite{jiang2023mistral}, Moondream~\cite{moondream2}, Qwen2.5-VL~\cite{qwen25vl_ollama}) across multiple model parameter sizes, via local hosting using the Ollama API (v0.11.10)~\cite{ollama}. We used pretrained weights without fine-tuning unless specified. Qwen2.5-VL was also evaluated using HuggingFace Transformers~\cite{qwen25vl_hf} for SFT, probing, and attention-head ablation to access model weights.
% (model ID: Qwen2.5-VL-7B-Instruct). 
% \MG{I would cite each respective technical report. Also, I'm not as familiar with "latest" notation, we usually put in the exact model name from the config files. So for example, llava-1.6, and then specify whether it's vicuna, mistral, or qwen underlying LM. Could be a difference of CVPR/ECCV notation vs ACL, so feel free to ignore if other papers do the same.}
Additional results from 12 recent VLMs, including both open-source on Hugging Face (e.g., InternVL3.5, Qwen3.6) and frontier closed-source models (e.g., GPT5.4, Gemini3.1, Claude Opus), are provided in the supplementary material.
% ~\cref{fig:frontier_scatter}.
\QZ{Add citations in Supplementary.}

Each stimulus-prompt pair was submitted in prompt-completion format, and requests were executed asynchronously using multiprocessing. From the responses, we parsed numerical slant estimates or curvature sign classes. We balanced the number of concave and convex test stimuli across experiments.
% Earlier experiments included confidence ratings and free-text reasoning to assess response variability under stochastic decoding. As results showed that predicted slant values were strongly anchored, subsequent analyses focus on slant estimates alone.
%
To capture trial-level variance, we repeated each query 10 times at the default temperature ($T=0.7$). As the observed variance was negligible, subsequent experiments used a lower temperature ($T=0.1$) with a single run per query for efficiency.

%% file: tables/prompt_natural_binary.tex
\begin{table}[b]
\centering
\caption{\textbf{Task prompt configurations.} Natural language style prompt variants and their component. \emph{Left:} Slant prediction as a regression task. \emph{Right:} Curvature sign estimation as a binary classification task.}
\resizebox{0.5\linewidth}{!}{
\begin{tabular}{l l}
\toprule
Prompt & Components Included \\
\midrule
0 & prompt\_minimal \\
1 & setup + task + format \\
2 & setup + task + format\_eg \\
3 & setup + task + format\_eg\_json \\
4 & setup + task + cues + format \\
5 & setup + task + cues + format\_eg \\
6 & setup + task + cues + format\_eg\_json \\
\bottomrule
\end{tabular}
}
\resizebox{0.4\linewidth}{!}{
\begin{tabular}{l l}
\toprule
Prompt & Components Included \\
\midrule
0 & prompt\_minimal \\
1 & setup + task + format \\
2 & setup + task + cues + format\\
\bottomrule
\end{tabular}
}
\label{tab:mainpaper_prompt_variants_natural}
\vspace{-0.5cm}
\end{table}

%% file: 4_results.tex
\vspace{-0.3cm}
\section{Results and Analysis}
\vspace{-0.15cm}

% We evaluate slant-from-texture under three conditions: zero-shot prompting, in-context learning with labeled examples, and supervised fine-tuning. 

\input{figures/results_natural_heatmap.tex}

%-------------------------------------------------------------------------------
\vspace{-0.15cm}
\subsection{Zero-shot slant estimation}
\vspace{-0.10cm}

\paragraph{Systematic anchoring.}
Across all models evaluated, zero-shot VLM predictions exhibit large errors in slant and curvature sign (\cref{fig:results_natural_heatmap}). Rather than varying smoothly with stimulus parameters, predicted slant values cluster around a small set of discrete anchor values, most commonly 0\degree and $\pm$45\degree, regardless of prompt detail. The percentages below each cell indicate the mode (most frequent number) as a proportion of all responses. In many cases, the same value appears in more than 50\% of responses, indicating anchoring rather than response noise.

We observed three qualitative response patterns:
\begin{itemize}[itemsep=1pt,topsep=0pt,leftmargin=12pt]
    \item \emph{Instruction compliance errors:} Models ignored format constraints, outputting text without numerical values or producing out-of-range values.
    \item \emph{Exemplar anchoring:} Prompts with examples induced ``example anchoring'' in some models, e.g., Moondream copying example prompt values verbatim (``-23\degree''), while other models were not affected by the specific values provided. 
    %; LLaVA-7B partially following)
    \item \emph{Zero-degree anchoring:} Models predicted flat surfaces (0\degree), sometimes accompanied by contradictory free-text reasoning in earlier runs (e.g., predicting flat while describing a steep slant).
\end{itemize}

\vspace{-0.2cm}
\paragraph{Anchoring persists across task framing, cues, prompt styles, and output formats.} We compute the mean absolute error (MAE) and variance of slant estimates across prompt detail variants, prompt styles and output formats, and task framing (\cref{fig:results_cmp}). We observe no significant improvement in VLM ability to predict slant or sign. Varying model temperature did not improve performance. Trial-level consistency analyses at temperature 0.7 showed stable anchoring across repetitions with low variance; averaging predictions did not improve performance. Anchoring is a systematic response pattern rather than stochastic variability. 
\input{figures/results.tex}

\vspace{-0.2cm}
\paragraph{In-context learning does not help.} In-context prompts do not significantly differ from other prompt families (\cref{fig:nattechcontext_cmp}): Median slant errors remain above 40\degree, curvature-sign accuracy stays near chance, and mode percentages remain high. ANOVA revealed no significant main effect of prompt type ($p{=}0.157$; \cref{tab:stats}). While specific anchor values sometimes shift across prompt conditions, the overall anchoring pattern persists.

\input{tables/results_stats.tex}
\input{figures/sft_cmp.tex}

\vspace{-0.2cm}
\paragraph{Summary.}
In zero-shot settings, VLM predictions collapse to discrete anchors, showing no systematic dependence on stimulus parameters, and are unaffected by prompt manipulations, including when given labeled examples for in-context learning. Model family and size modulate variability but do not alter the underlying anchoring pattern. Model performance is uniformly poor, making differences between models difficult to interpret.
% \MG{More concrete insight; see comment.}
% \MG{It would be nice to give concrete insights into each model family. The only model explicitly mentioned is LLaVA-7B, but it would be nice to say, for example, Qwen family defaults to X while Gemma and Mistral defualt to Y. Then maybe make a comment on their different fine-tuning paradigms from their respective technical reports (optional). Similarly, dive deeper into whether model size makes a difference, for example for qwen and llama family there is no difference in json3 but for llavas 13b does -45 whereas 7B predicts 45. }
% \QZ{I am uncertain if the anchoring values mean anything useful, as I see different values across Ollama and huggingface for the same model. Will look into this later.}
%
This behavior is consistent with evidence that LLM-style decoders exhibit anchoring effects and produce coarse numeric estimates that concentrate on a small set of preferred values, including round-number biases~\cite{tseng2025streetmath, lou2026anchoring}; related work also identifies prompt-copying in VLMs specifically~\cite{rudman2026mechanisms}. A complementary explanation is that round numbers are disproportionately frequent in natural language use, making a few canonical numerals strong attractors when models generate numbers as text~\cite{woodin2024large}. \QZ{Read woodin2024large again: what exactly did they say about round numbers - scattered around a few values, or continuous integers?}

%-------------------------------------------------------------------------------
\vspace{-0.15cm}
\subsection{Supervised fine-tuning (SFT)}
\label{sec:results_sft}
\vspace{-0.10cm}
%Next, we next evaluated whether supervised fine-tuning alters VLM behavior on the slant-from-texture task. 
We fine-tuned Qwen2.5-VL-3B using LoRA with a masked token loss on labeled slant and curvature data, updating the vision-language fusion layers (q\_proj and v\_proj).\footnote{LoRA was restricted to q/v\_proj following standard PEFT practice~\cite{peft}; the regression-head result in \cref{sec:results_vit} confirms this choice does not determine the bottleneck as frozen post-projector features are already linearly decodable for slant.} Qwen2.5-VL-3B was chosen for its consistent anchoring patterns across regression and classification settings and its moderate size for efficient fine-tuning.

% Fine-tuning directly modifies how fused vision-language representations are translated into textual outputs, yet does not affect the inductive bias in vision and language modules, respectively. 
% \input{figures/sft_diagram.tex}

%For fine-tuning analysis, we intentionally reverse the train/test split used in zero-shot experiments, using the larger 2000-image set for evaluation. This provides denser sampling of each slant-FOV condition (10 images per cell), enabling more reliable analysis of curvature-dependent error patterns. This design choice does not affect the qualitative anchoring patterns reported below. We use this reversed split for analytical resolution, not to evaluate generalization; the denser sampling enables reliable curvature-dependent error analysis.
% \MG{I'm a little lost here, why is the last sentence needed?}
% \JT{Agreed; let's simplify it.}

\vspace{-0.2cm}
\paragraph{Improved slant estimation anda finer-scale anchoring.}
SFT substantially improves slant estimation and curvature-sign discrimination (\cref{fig:results_sft_cmp}). Mean slant error decreases monotonically with training (MAE: $48.3\degree\rightarrow18.9\degree\rightarrow5.6\degree\rightarrow4.7\degree$ at $0/3/10/100$ epochs), and the predictions expand from a few discrete anchors toward a more continuous range (peak anchor count: $208\rightarrow114\rightarrow30\rightarrow10$; distinct predicted values: $3\rightarrow38\rightarrow70\rightarrow171$). 
Paired statistical tests confirm significant improvement at every checkpoint (test set, $df=399$; e.g. $100$ epochs: $t=20.8$, $p=9.9\times10^{-66}$). 
Curvature-sign accuracy rises from chance (50\%) to 96.8\%, comparable to unsupervised CNNs (96.4\%)~\cite{wang2023human} and above human performance (86\%)~\cite{todd2005effects}, though the underlying error distributions differ qualitatively.
Despite these gains, SFT reduces but does not eliminate anchoring.
Compared to before SFT, anchoring persists at a progressively finer scale through fine-tuning and is distributed around the ground-truth slant (diagonal line), forming horizontal bands (\cref{fig:results_sft_scatter}). The most frequent anchors are indicated by the horizontal gray dashed lines, with the count on the right side. Even after 100 epochs the predictions do not fluctuate smoothly about the true slant.

\input{figures/sft_scatter.tex}

\vspace{-0.2cm}
\paragraph{Convex-concave asymmetry and FOV dependent effects.}
We analyze curvature sign errors as a function of field of view and optical slant (~\cref{fig:results_sft_heatmaps}, sign-flip frequency).
Shape discrimination errors are elevated at small FOVs and decrease systematically with larger FOV and optical slant. These errors are \emph{asymmetric} across curvature sign: concave stimuli show near-zero flip rates at large fields of view, whereas convex stimuli retain moderate error rates (0.3--0.4) across optical slant. This pattern varies in the same direction as human behavior (B1), though it is more pronounced.
% The curvature-sign judgment errors change with FOVs, larger at low FOVs (\cref{fig:results_sft_scatter}), which is consistent with the FOV dependent error patterns observed in human judgments.
We plot the accuracy of curvature-sign prediction as a function of optical slant and FOV and compare with CNN results and human study in \cref{fig:sft_acc_vs_slant}. They show similar FOV-dependent error patterns.

\paragraph{Summary.}
SFT effectively reduces anchoring but does not eliminate it. Predictions become monotonically related to ground-truth slant and curvature-sign discrimination improves and becomes closer to past human and CNN studies. This suggests that while SFT can steer model behavior, it does not fully override the anchoring pattern.

\input{figures/sft_convex_concave.tex}
\input{figures/sft_acc.tex}
\vspace{-0.2cm}

% \paragraph{SFT on other VLMs.}

% \QZ{Chameleon SFT works similarly to the others: format prompt+completion, LoRA on q_proj/v_proj of the decoder, causal-LM loss on the answer.
% The only difference from llava/paligemma: there's no model.vision_tower / model.multi_modal_projector to freeze. The VQ tokenizer (model.model.vqmodel) is the "vision" part and is already frozen (not a LoRA target). So Chameleon's registry entry sets freeze=None and uses ChameleonProcessor in code; everything else is shared.}

%% file: figures/results_natural_heatmap.tex
\begin{figure}[t]
\centering
    \includegraphics[height=5.0cm]{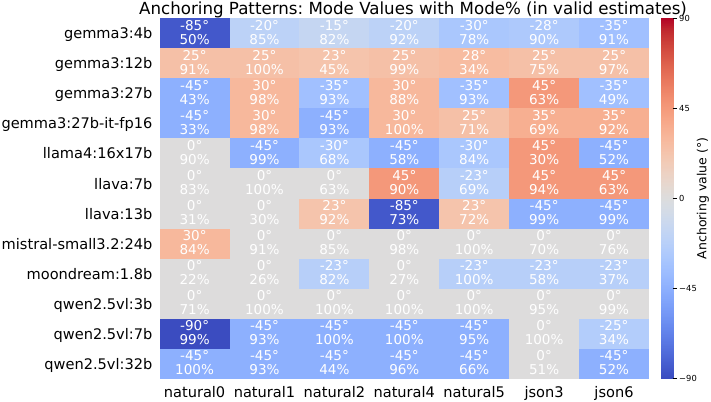}
    \includegraphics[height=5.0cm]{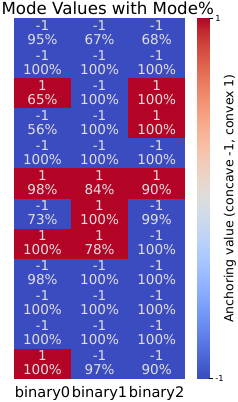}
\vspace{-0.3cm}
\caption{\textbf{VLMs anchor significantly on slant angle and sign prediction.} Anchoring patterns for natural language prompts in regression task (left) and binary classification task (right). The heatmap shows distributions of estimated slant across VLMs (rows) and natural prompt variants (columns). All models collapse to a few discrete anchors (e.g., 0\degree, ±45\degree) regardless of prompt detail, indicating that prompt variation does not mitigate anchoring. Prompt indices (e.g., ``natural0'') and their components are in~\cref{tab:mainpaper_prompt_variants_natural}, including both with and without JSON output constraints.
}\label{fig:results_natural_heatmap}
% \QZ{Will replace the latest tag with exact model version-size tag and update the figures. Ollama models have shorter names, and are not always updated to the latest version.}
\vspace{-0.6cm}
\end{figure}

%% file: figures/results.tex
\begin{figure}[b]
\centering
\vspace{-0.2cm} 
%% Row (a)
\begin{minipage}[t]{0.48\linewidth}
    \vspace{0pt}
    \centering
    \includegraphics[width=1.0\linewidth,clip,trim={0cm 0cm 0cm 0.3cm}]{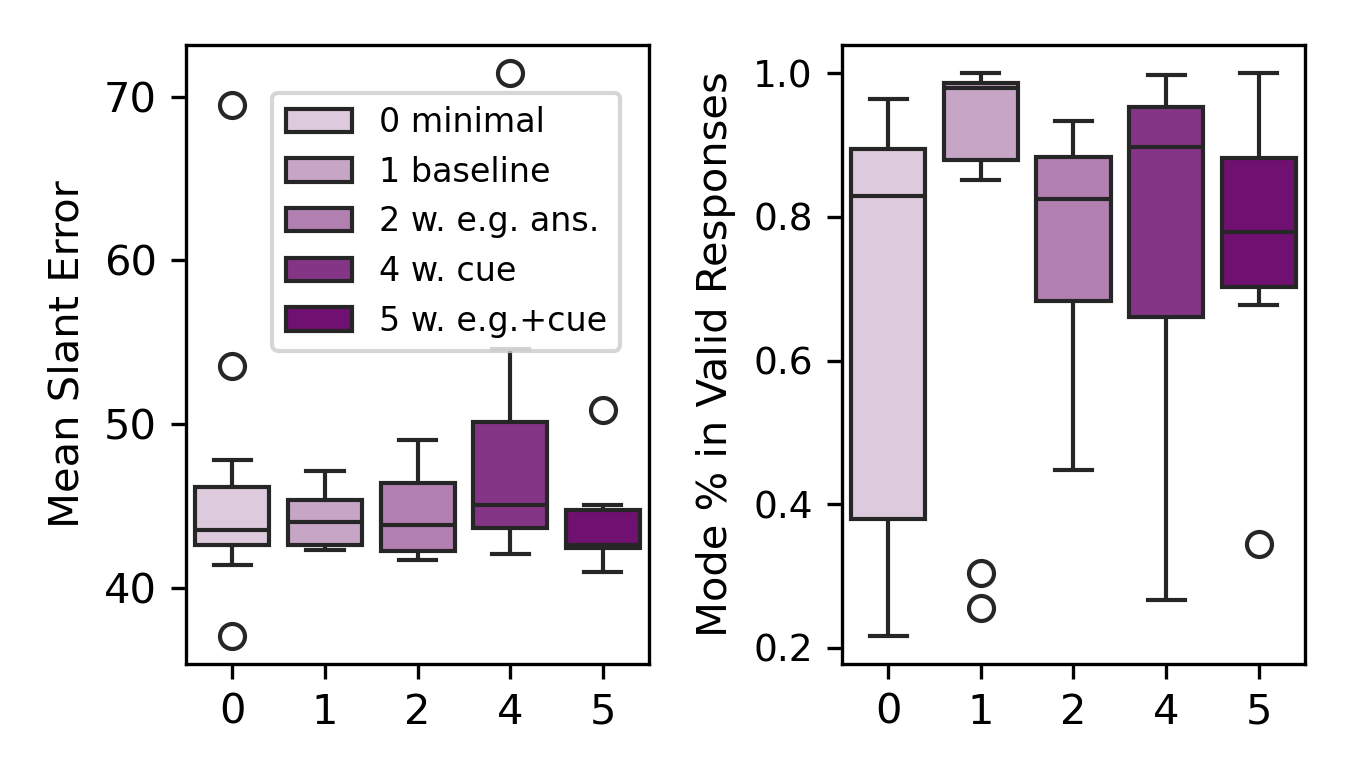}
    \vspace{-0.6cm}
    \subcaption{Prompt detail (natural style)}\label{fig:results_natural_prompt_detail}
\end{minipage}%
\hfill
\begin{minipage}[t]{0.48\linewidth}
    \vspace{0pt}
    \scriptsize
    Varying prompt detail does not reduce anchoring. The anchor mode value changed for some models, but median errors are above 40\degree~and anchoring dominates. Prompts emphasizing geometric cues yielded no consistent improvement (e.g., descriptions of texture scaling as included in prompts 4 and 5).
\end{minipage}

\vspace{0.1cm}

%% Row (b)
\begin{minipage}[t]{0.7\linewidth}
    \vspace{0pt}
    \centering
    \includegraphics[width=1.0\linewidth,clip,trim={0cm 0cm 0cm 0.3cm}]{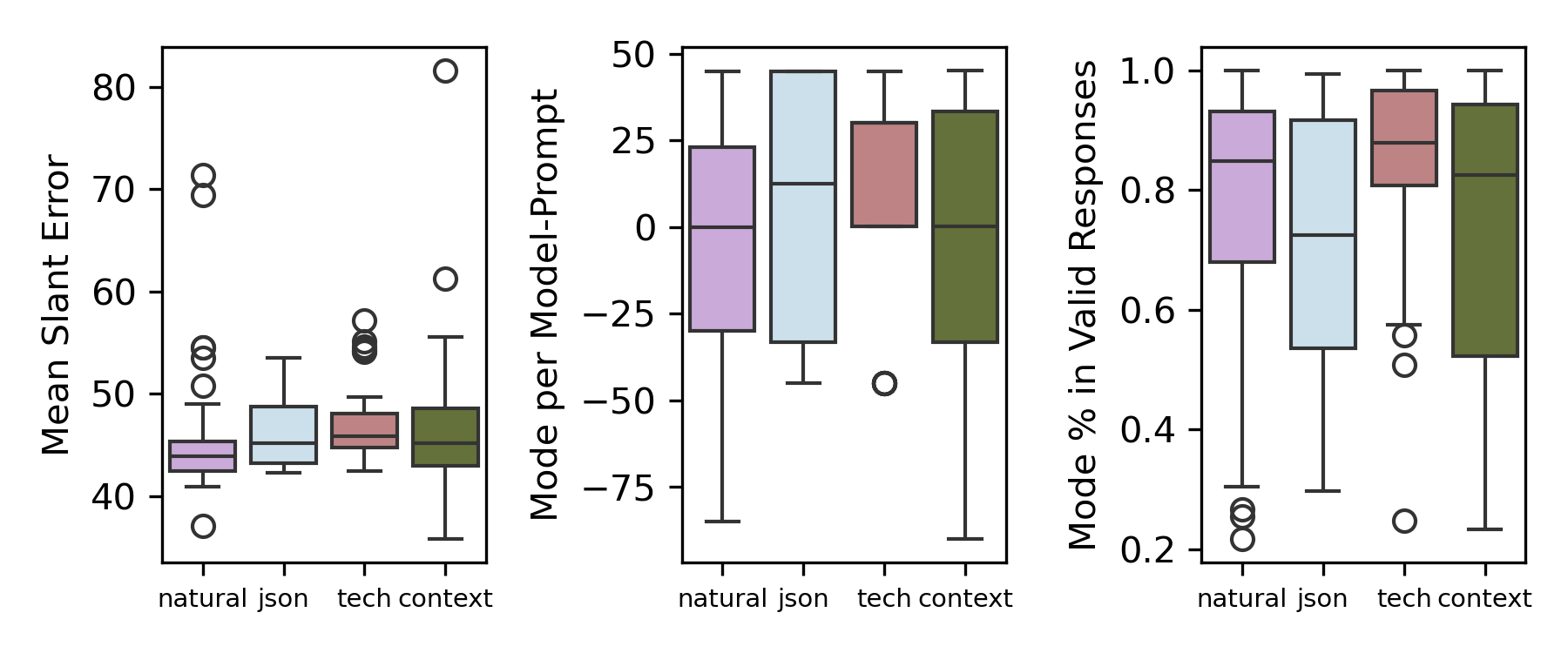}
    \vspace{-0.6cm}
    \subcaption{Prompt style and output format}\label{fig:nattechcontext_cmp}
\end{minipage}%
\hfill
\begin{minipage}[t]{0.28\linewidth}
    \vspace{0pt}
    \scriptsize
    Prompt style (natural vs.\ technical), output format (JSON), and in-context examples show no significant variation (two-way ANOVA on main effect of prompt family on slant error; \cref{tab:stats}). JSON formatting does not consistently reduce example anchoring in any models but does exhibit increased overall variance, i.e., mode percentages decrease.
\end{minipage}

\vspace{0.1cm}

%% Row (c)
\begin{minipage}[t]{0.27\linewidth}
    \vspace{0pt}
    \centering
    \includegraphics[width=1.0\linewidth,clip,trim={0cm 0cm 0cm 0.3cm}]{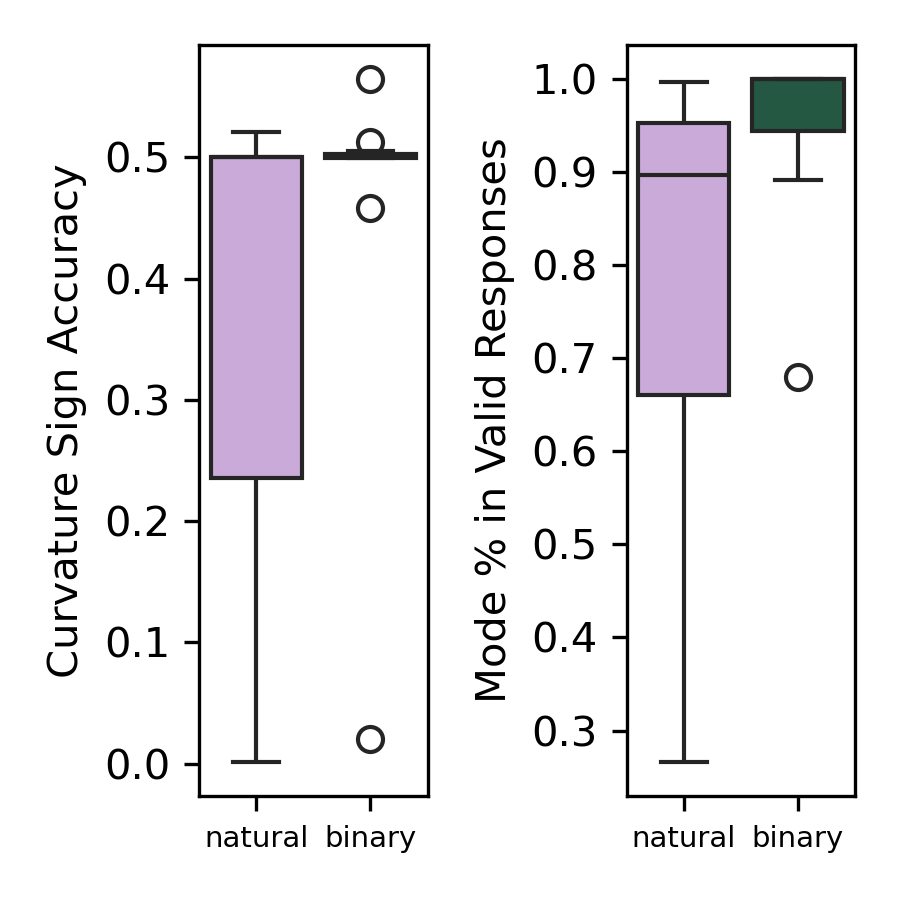}
    \vspace{-0.6cm}
    \subcaption{Task framing}\label{fig:binary_cmp}
\end{minipage}%
\hfill
\begin{minipage}[t]{0.68\linewidth}
    \vspace{0pt}
    \scriptsize
    Curvature sign prediction as convex vs.\ concave (classification, ``binary'' in plot) yielded no improvements over slant prediction (regression, ``natural'' in plot), with accuracies around 0.5 (chance) and mode percentage increasing to around 90\%. This trend is consistent across models.
\end{minipage}

\vspace{-0.2cm}
\caption{Effects of prompt and task framing as boxplots of 95\% Confidence Intervals.}\label{fig:results_cmp}
\end{figure}

% \begin{figure}[b]
% \begin{center}
%    \includegraphics[width=1.0\linewidth]{images/natural/prompt_detail.png}
% \end{center}
% \caption{Comparison of slant MAE and mode by detail level in prompts for the natural language prompts. }\label{fig:results_natural_prompt_detail}
% \end{figure}

% \begin{subfigure}[t]{0.45\linewidth}
%     \centering
%     \includegraphics[width=1.0\linewidth]{images/incontext/incontext_cmp.png}
%     \vspace{-0.6cm}
%     \caption{Comparison of slant prediction error of VLM estimations from natural and in-context learning.}\label{fig:sft_cmp}
% \end{subfigure}

% \begin{subfigure}[t]{0.45\linewidth}
%     \centering
%     \includegraphics[width=1.0\linewidth]{images/incontext/incontext_cmp.png}
%     \vspace{-0.6cm}
%     \caption{Comparison of slant prediction error of VLM estimations from natural and in-context learning.}\label{fig:incontext_cmp}
% \end{subfigure}

% \begin{subfigure}[t]{0.45\linewidth}
%     \centering
%     \includegraphics[width=1.0\linewidth]{images/natural/json.png}
%     \vspace{-0.6cm}
%     \caption{Averaged slant error across models for prompts requiring answers in plain text vs.\ JSON format.}
%     \label{fig:natural_json}
% \end{subfigure}
% \hfill
% \begin{subfigure}[t]{0.45\linewidth}
%     \centering
%     \includegraphics[width=1.0\linewidth]{images/technical/technical_cmp.png}
%     \vspace{-0.6cm}
%     \caption{Mean slant error averaged across models for prompt styles: natural language vs.\ technical.}
%     \label{fig:technical_cmp}
% \end{subfigure}==

%% file: tables/results_stats.tex
\begin{table}[t]
\begin{minipage}[t]{0.40\linewidth}
    \vspace{2pt}
    \centering
    \resizebox{\linewidth}{!}{
    \begin{tabular}{l r r r r}
    \toprule
    Source & sum\_sq & df & F & p-value \\
    \midrule
    C(model)                  & 753.19   & 6  & 4.81  & 0.000291 \\
    C(prompt\_type)           & 139.32   & 3  & 1.78  & 0.157403 \\
    % C(model):C(prompt\_type)  & 1230.32  & 18 & 2.62  & 0.001634 \\
    Interaction  & 1230.32  & 18 & 2.62  & 0.001634 \\
    Residual                  & 2192.87  & 84 & --    & --       \\
    \bottomrule
    \end{tabular}
    }
\end{minipage}%
\hfill
\begin{minipage}[t]{0.58\linewidth}
    \vspace{0pt}
    \caption{Two-way ANOVA (with interaction) of slant error mean for different models
    %(Gemma3, Llama4, LLaVA, Moondream) 
    and prompts.
    %(natural, JSON, technical, in-context). 
    Prompt type alone does not have a significant effect but depends on the model, e.g., one model might vary with JSON while another does not. Overall, there is no consistent main effect.}\label{tab:stats}
\end{minipage}
\vspace{-0.6cm}
\end{table}

%% file: figures/sft_cmp.tex
\begin{figure}[t]
\begin{minipage}[t]{0.40\linewidth}
    \vspace{0pt}
    \centering
    \includegraphics[width=1.0\linewidth]{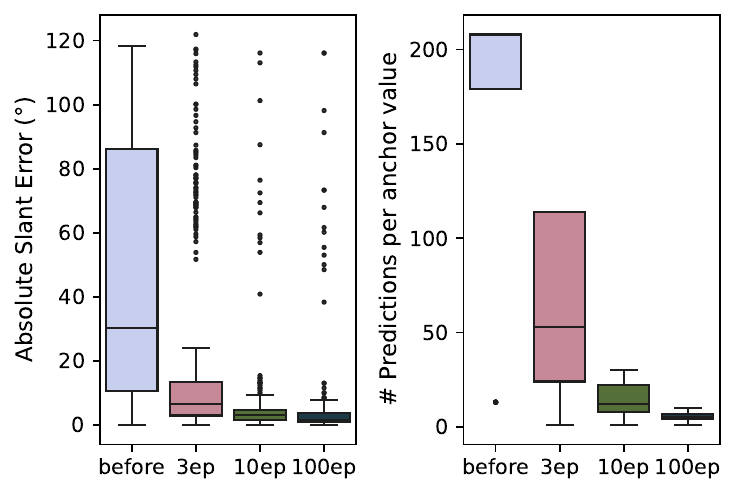}
\end{minipage}%
\hfill
\begin{minipage}[t]{0.59\linewidth}
    \vspace{0pt}
    \caption{\textbf{Fine-tuning partially helps.} Qwen2.5-VL before and after supervised fine-tuning (SFT). \emph{Left:} absolute slant error. \emph{Right:} number of predictions per anchor value. Slant error decreases and anchoring reduces after SFT, but even after 100 epochs, anchoring persists.}\label{fig:results_sft_cmp}
\end{minipage}
\vspace{-0.6cm}
\end{figure}

%% file: figures/sft_scatter.tex
\begin{figure}[t]
\centering
    \includegraphics[width=1.0\linewidth]{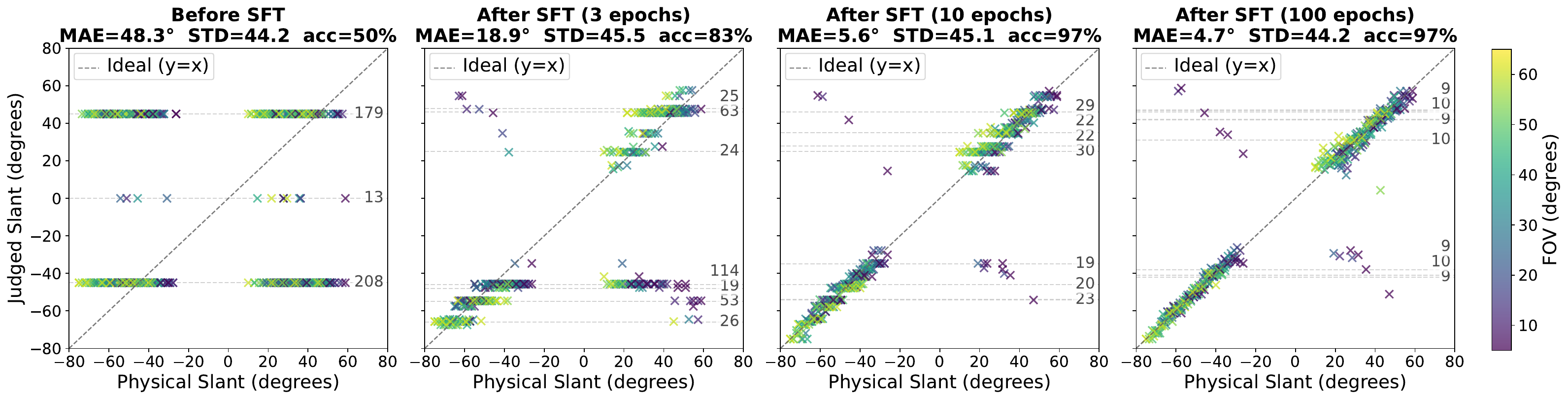}
    \vspace{-0.6cm}
    \caption{\textbf{SFT improves slant estimation but anchoring persists.} Judged vs.\ ground-truth slant on held-out test set. Left to right: before SFT and after 3, 10, and 100 epochs. Gray dashed lines mark the dominant predicted (anchor) values; the number to the right of each horizontal line is how many of the 400 test estimations fall on it.
    % VLMs before supervised fine-tuning show anchoring; bar chart of slant error distribution; VLMs with 10 epochs of SFT; VLMs with 100 epochs of SFT. After SFT, VLMs show improved slant prediction, but persistently estimate incorrect curvature sign, especially for low FOVs.
    }
    \label{fig:results_sft_scatter}
\vspace{-0.6cm}
\end{figure}

%% file: figures/sft_convex_concave.tex
\begin{figure}[t]
\begin{minipage}[t]{0.50\linewidth}
    \vspace{0pt}
    \centering
    \includegraphics[width=1.0\linewidth]{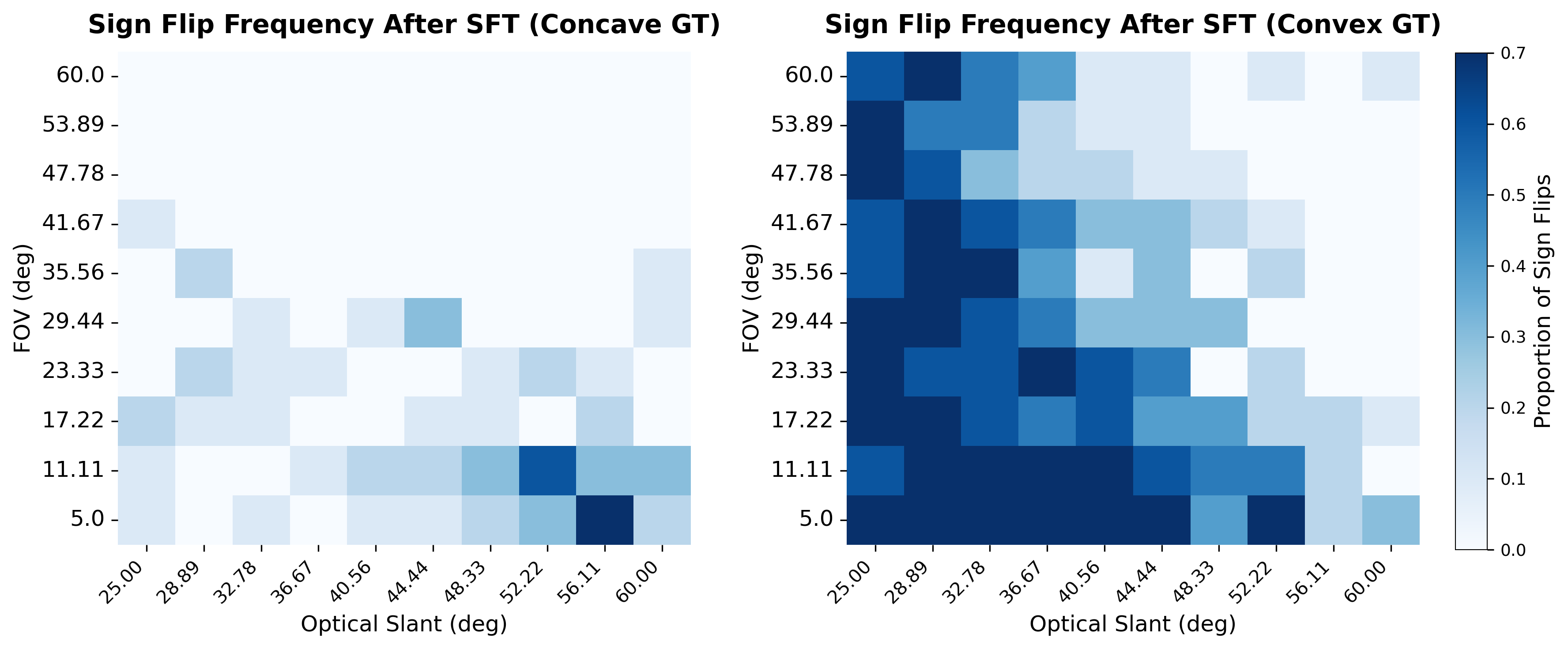}
\end{minipage}
\hfill
\begin{minipage}[t]{0.47\linewidth}
    \vspace{0pt}
    \caption{SFT error analysis grid by slant angle and FOV, showing frequency of sign flips in predictions. Larger FOVs and larger optical slant images induced fewer errors (sign flips); no curvature sign errors for concave FOV$>45$\degree; convex-to-concave misjudge is more frequent than vice versa.}
    \label{fig:results_sft_heatmaps}
% \QZ{Accuracy is the convention, rather than error rate. Re-plot?}
\end{minipage}
\vspace{-0.5cm}
\end{figure}

% \begin{figure}[t]
% \begin{minipage}[t]{0.4\linewidth}
%     \vspace{0pt}
%     \centering
%     \includegraphics[width=1.0\linewidth]{images/sft/sft_cmp.png}
% \end{minipage}%
% \hfill
% \begin{minipage}[t]{0.58\linewidth}
%     \vspace{0pt}
%     \caption{\textbf{Fine tuning only partially helps.} Qwen2.5-VL before and after supervised fine-tuning: slant error decreases and anchoring weakens after SFT, but many outliers persist with high error.}\label{fig:results_sft_cmp_fig}
%     \label{fig:results_sft_cmp}
% \end{minipage}
% \vspace{-0.6cm}
% \end{figure}

%% file: figures/sft_acc.tex
\begin{figure}[t]
\centering
    \includegraphics[width=0.28\linewidth]{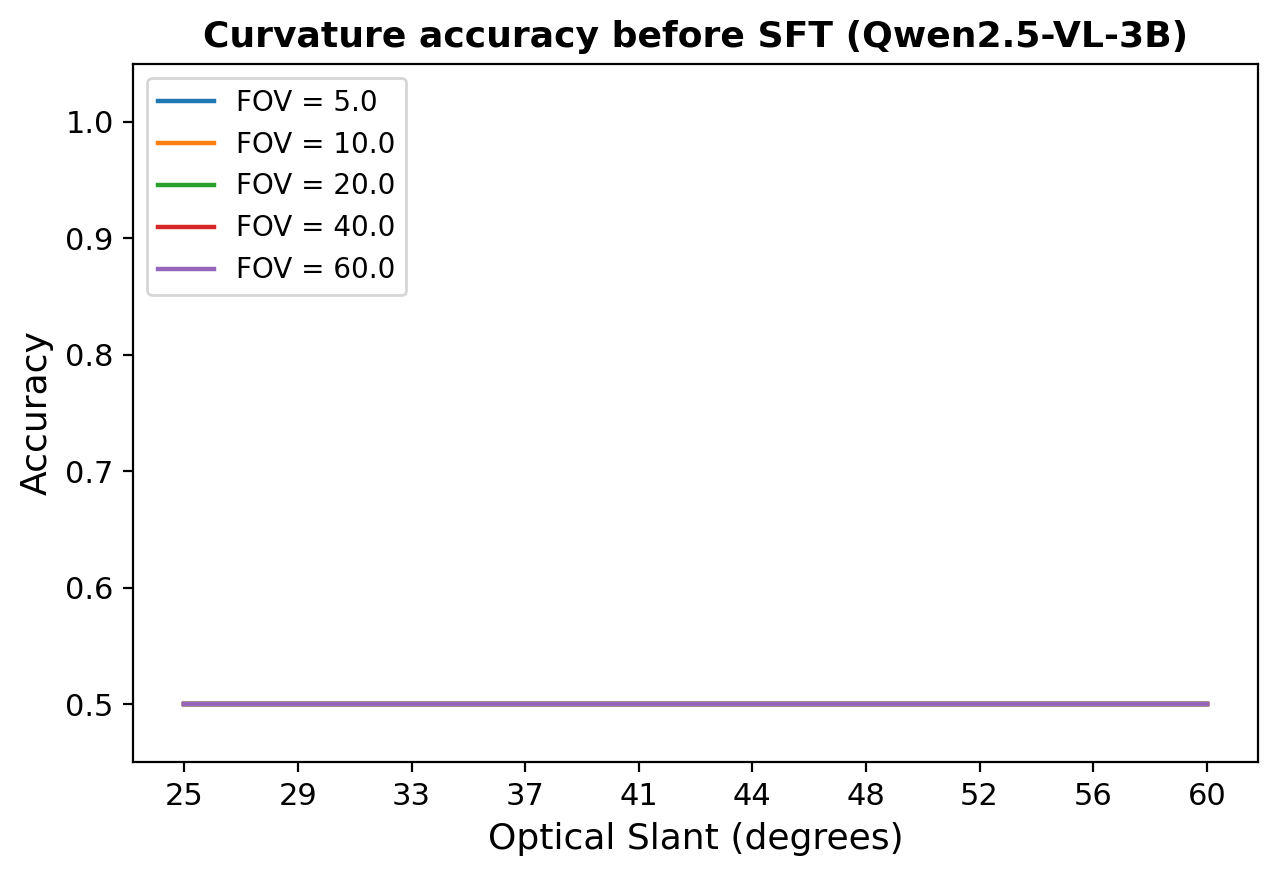}
    \includegraphics[width=0.28\linewidth]{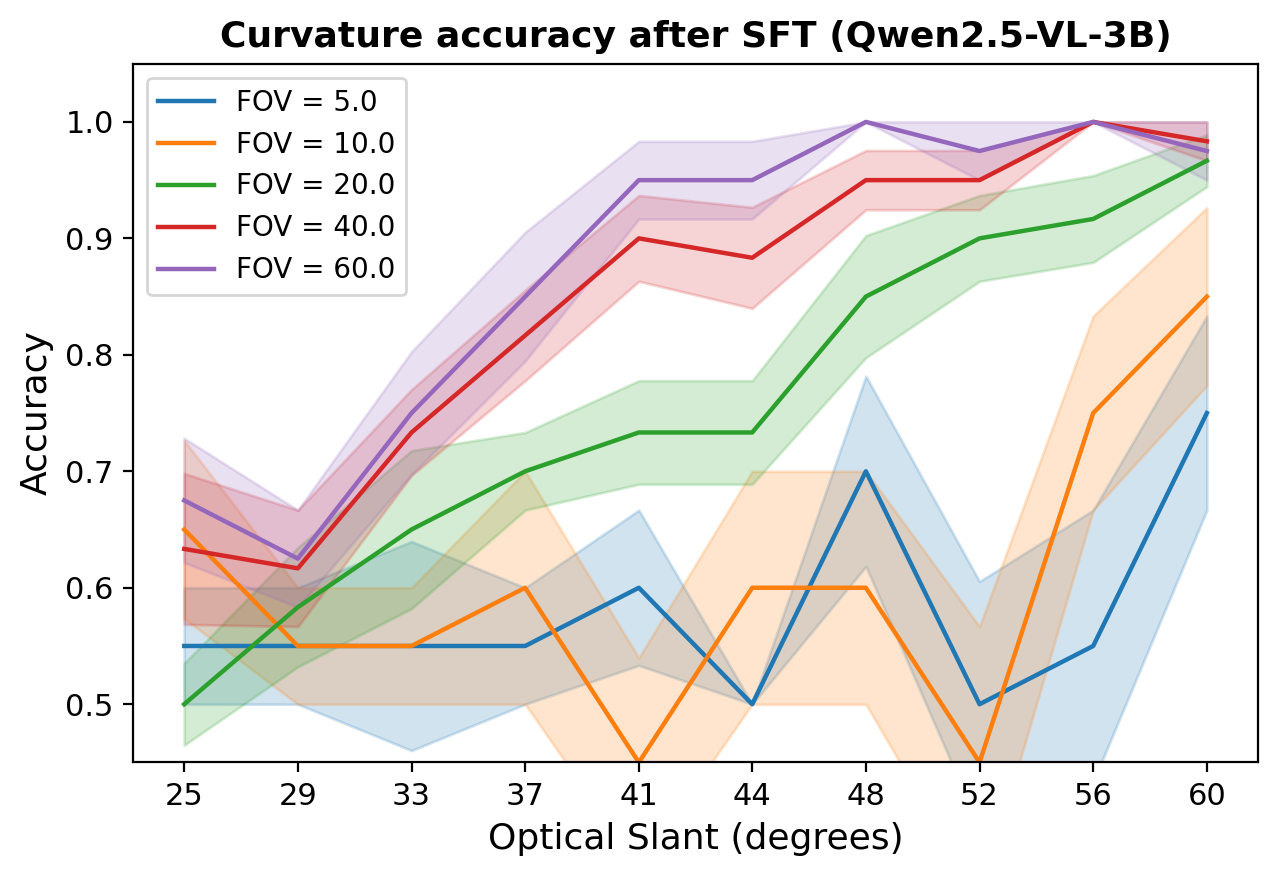}
    \includegraphics[width=0.23\linewidth]{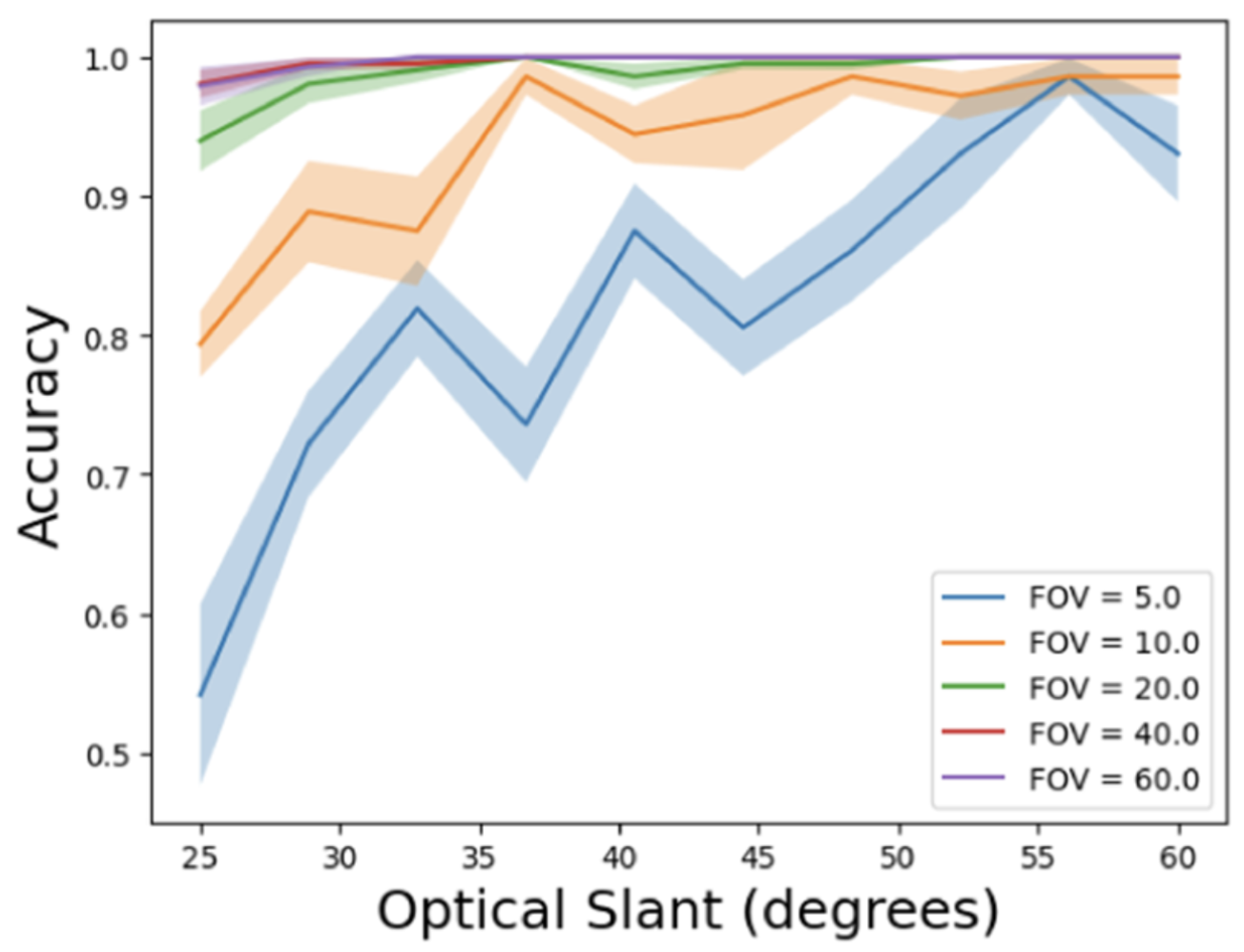}
    \includegraphics[width=0.19\linewidth]{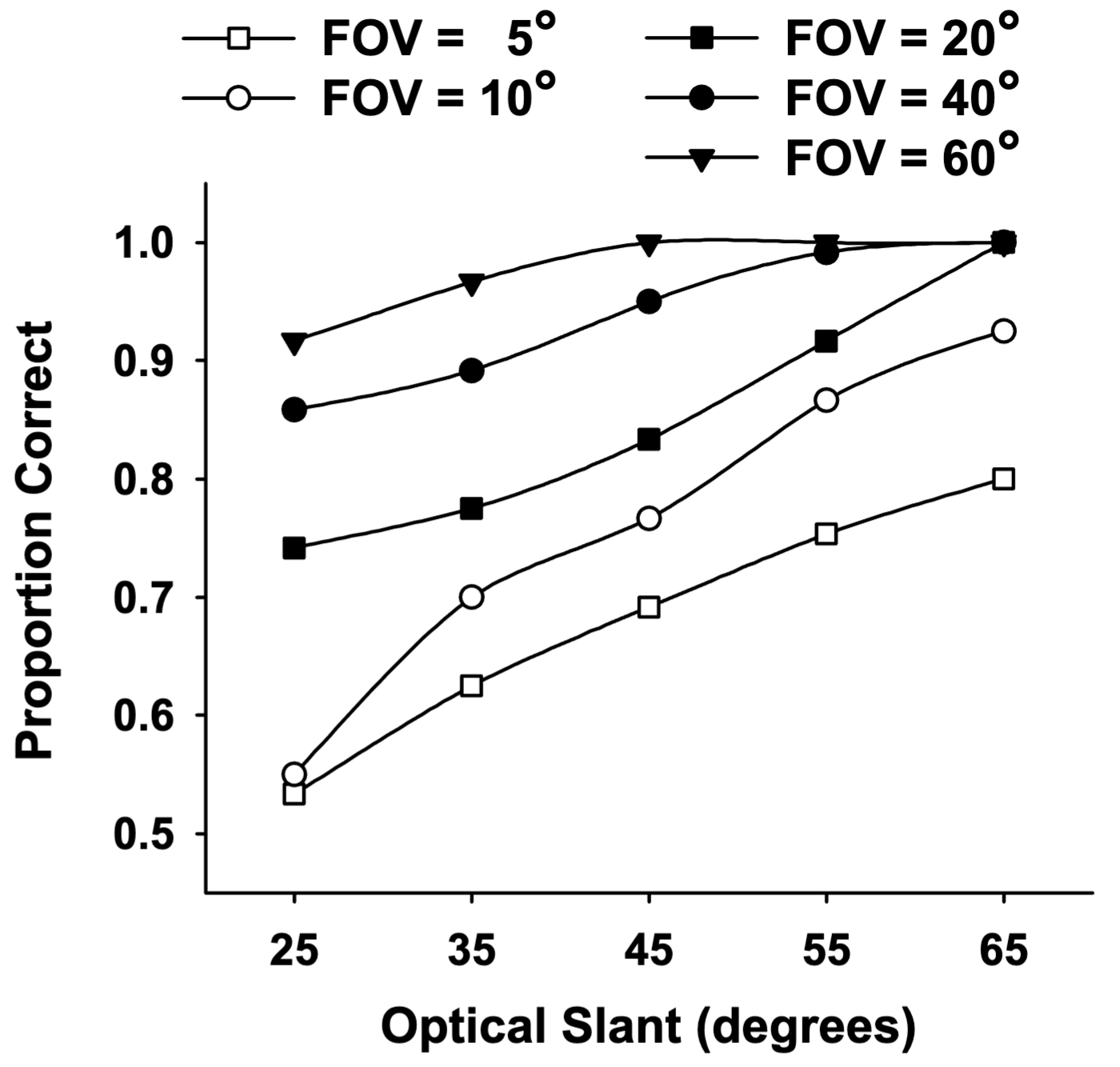}
\vspace{-0.3cm}
\caption{VLM, CNN, and human curvature sign judgment accuracy versus optical slant per FOV, shown in matching axes for direct comparison. 
\emph{Left:} Performance of VLM Qwen2.5-VL-3B before supervised fine-tuning (SFT). 
\emph{Center left:} After 3 epochs of SFT, Qwen2.5-VL shows improved accuracy with increasing FOV and optical slant.
\emph{Center right:} Unsupervised CNN judgments from Wang et al.~\cite{wang2023human}.
\emph{Right:} Human judgments from Todd et al.~\cite{todd2005effects}.
\QZ{Align y-axis of subplots}
}\label{fig:sft_acc_vs_slant}
\vspace{-0.6cm}
\end{figure}

%% file: 4_results2_vit.tex
%-------------------------------------------------------------------------------
% \vspace{-0.25cm}
\subsection{Probing the Vision Module}
\label{sec:results_vit}
% \vspace{-0.15cm}

% VLMs fail at slant-from-texture tasks, producing stereotyped numerical responses (response anchoring) that do not co-vary with stimulus geometry.
%Does the vision encoder already contains geometric information - 
Pre-trained VLMs are trained on millions, often billions, of image–caption pairs, from which a complex structure is thought to emerge in their representations. Given that VLMs perform reasonably when directly trained on the stimuli and task, we use linear probing~\cite{alain2016understanding} to \emph{localize} where correct behavior fails to emerge. The key question is whether zero-shot VLM failures arise from incorrectly encoded geometric information or from a deficit in language readout.

%-------------------------------
% \vspace{-0.1cm}
\paragraph{The ViT architecture.}
We analyse the vision encoder of Qwen2.5-VL-3B, a 32-block transformer
(width 1280, 16 heads, SwiGLU feed-forward of width 3420) using RMSNorm
and 2D rotary position embeddings. 
Attention is windowed ($112\times 112$ px) in all but four blocks, which use full attention.
Each stimulus is tokenised by a Conv3D patch embedding with kernel
$(2, 14, 14)$; still images are duplicated along the temporal axis, so a
$224\times 224$ stimulus yields a $16\times 16$ grid of 1280-D patch tokens. 
A patch merger then pools each $2\times 2$ spatial neighbourhood
through a two-layer MLP, reducing the token count fourfold and
projecting to the 2048-D language-model width, giving 64 vision tokens
that are concatenated with the text tokens in the decoder's input sequence.
We probe these 64 post-merger tokens with the backbone frozen and no
fine-tuning, mean-pooling across tokens to obtain one 2048-D vector per
stimulus. Since the feature dimensionality exceeds the sample count
($dim = 2048$, $\#samples = 400$), we fit ridge regression for all VLM probes with $\alpha = 1.0$; all reported $R^2$ values are computed on held-out test data.

\input{figures/vit_vlm_layerwise.tex}
%-------------------------------
\vspace{-0.1cm}
\paragraph{Post-merger probing results.}
We evaluate regression fits using the coefficient of determination ($R^2$), which measures the proportion of variance in the target variable (PS, OS, FOV) explained by a probe trained on the latent tokens. On the test set, $R^2{=}1$ indicates perfect prediction, while $R^2{=}0$ indicates performance equivalent to predicting the mean. Ridge probes on mean-pooled VLM vision tokens achieve physical slant $R^2{=}0.828$, optical slant $R^2{=}0.988$, FOV $R^2{=}0.884$, and curvature accuracy $=0.96$, indicating that geometric information is strongly encoded in the vision representations. These results place the VLM vision encoder in the same performance regime as a pretrained ViT-MAE-86M~\cite{he2022masked} (PS $0.856$, OS $0.997$, FOV $R^2{=}0.874$), suggesting VLM vision features capture geometry at a level comparable to this standalone vision backbone.

% CLIP is a VLM.
% DINOv3 (Base + Small 22\M ), 
% ViT-MAE FOV=0.874 PS=0.856 OS=0.997

%-------------------------------
\vspace{-0.1cm}
\paragraph{Layer-wise probing.}
To trace how geometric information flows through the vision encoder, we extract mean-pooled features from every intermediate layer using forward hooks: the patch embedding output, all 32 transformer blocks, and the post-merger output.
Applying ridge regression probes independently at each layer reveals a clear progression (\cref{fig:vlm_layerwise}).
The raw patch embedding contains almost no geometric information (OS $R^2=0.031$, all others near zero; curvature accuracy $=0.565 \approx$ chance), indicating that the Conv3D projection captures little beyond local texture statistics.
However, by the first transformer block (layer~0), optical slant is already near ceiling ($R^2 = 0.992$), confirming that it is a low-level image property recoverable from local texture gradients after minimal processing.

In contrast, physical slant, FOV, and curvature require progressively deeper processing:
FOV $R^2$ rises from $0.474$ (layer~0) to $0.920$ (layer~18);
physical slant from $0.329$ to $0.879$;
curvature accuracy from $0.690$ to $0.995$.
The correlation of all three variables peaks around layers 17--19 (the 56th--59th percentile of the 32-layer vision encoder) and then \emph{declines} in the final layers: by layer~31, FOV drops to $0.844$, physical slant to $0.769$, and curvature sign accuracy to $0.96$.
The patch merger partially recovers from this late decline (FOV $= 0.884$, PS $= 0.828$), likely because its spatial $2{\times}2$ merging reintroduces local structure that the late layers had redistributed.

This late-layer decline is consistent with the view that the final transformer blocks specialize in the downstream language modeling objective, i.e., reorganizing features for cross-modal projection at the cost of some geometric accessibility.
The practical implication is that middle layers of the vision encoder (around layer~18) contain the richest geometric signal, not the final output.

%-------------------------------
\paragraph{Generalization across vision encoders.}
\CC{To test whether the mid-/late-layer geometric-encoding peak is specific to Qwen2.5-VL's vision tower, we replicate the analysis on three additional VLMs with different vision encoders and language models: LLaVA-1.5-7B~\cite{llava_hf}, PaliGemma-3B~\cite{beyer2024paligemma}, and Chameleon-7B~\cite{chameleon7b} \footnote{As an architecture-diversity test, the sizes are each architecture's canonical released size, not a controlled choice: LLaVA-1.5 has no 3B and its smallest is 7B; PaliGemma is 3B only. A true size-controlled SFT comparison would be a separate experiment.} (\cref{tab:layerwise-probe-peaks}). For each, we hook the per-layer outputs of the vision tower (including the projector), mean-pool the patch tokens, and fit ridge regressions (physical slant, optical slant, FOV) and a logistic probe (curvature sign).}

\input{tables/vlm_vision_module_cmp.tex}

\CC{As shown in~\cref{fig:layerwise_generalization}, we found that geometric encoding generalizes across vision-encoder architectures. Across all four models, the layer-wise $R^2$ profile is qualitatively similar: optical slant saturates at $R^2$ $\geq$ 0.997 by the first encoder layer; physical slant climbs gradually to a peak of 0.88--0.93 at 60--80\% normalized layer depth; curvature accuracy reaches 0.99--1.00; and the projector preserves $\geq$ 95\% of the late-layer peak. Our finding holds across vision towers.}

\CC{One model of interest is Chameleon: the only early-fusion model, where vision and text tokens flow through the same transformer. Layer 31, the final layer feeding into the language model (LM) head, shows a deep drop in geometric correlation: PS $R^2$ = 0.065, OS $R^2$ = 0.033, FOV $R^2$ = 0.005, curvature accuracy = 0.552 (chance). The geometric information present at layers 12--30 ($R^2 > 0.9$) is transformed away in the final layer toward the language-output distribution. This supports our claim that encoding is preserved deep in the network, but the final language-generation step discards it.}

% \QZ{For chameleon, we do not probe the vision encoder because there isn't one. We hook the unified decoder's 32 layers (model.model.layers) and mean-pools hidden states over only the image-token positions (the 1024 VQ-VAE codes after the <image> marker, extract_chameleon_layerwise_may11.py:101-116).
% So "vision features" for Chameleon = decoder hidden states restricted to image tokens. It is "conceptually different" from other VLMs.}

%-------------------------------
\paragraph{Readout test.}
\input{figures/vlm_frozen_vision_regression.tex}

\CC{
To test whether this is a true readout problem, we regress physical slant on the LM-input vision tokens, asking whether they contain the necessary geometric information that the language model cannot translate into a continuous estimate. We mean-pooled the frozen post-projector tokens and trained a single ridge-regression head on the 2000-image training split, with no further tuning. As shown in~\cref{fig:token_regression}, on the held-out 400-image test set, this single linear projection predicts physical slant at $R^2$ = 0.828 (Qwen2.5-VL-3B; MAE 14.5\degree), $R^2$ = 0.890 (LLaVA-1.5-7B; MAE 10.2\degree), $R^2$ = 0.893 (PaliGemma-3B; MAE 10.5\degree), and produces 333--354 of 400 unique predicted angles, i.e., predictions are essentially continuous.
This shows directly the readout bottleneck, as the geometry is decodable but the language head discards it.
}

% \QZ{Chameleon-7B proving result missing; different API.}

\input{figures/lm_ablation.tex}

%% file: figures/vit_vlm_layerwise.tex
\begin{figure}[t]
    \centering
    \includegraphics[width=0.7\linewidth]{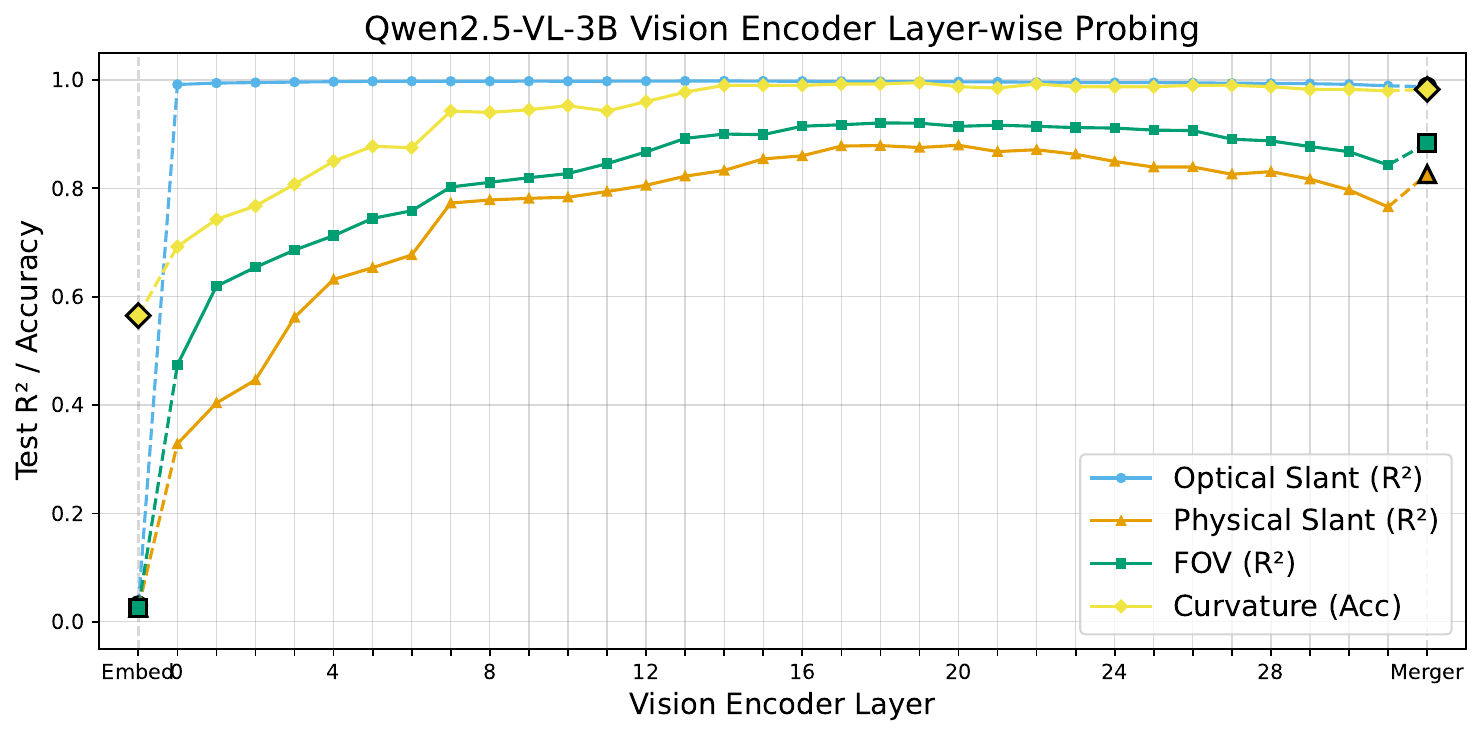}
    \vspace{-0.4cm}
    \caption{Layer-wise linear probe performance across the 32-layer Qwen2.5-VL-3B vision encoder. Optical slant is near ceiling from the first transformer layer ($R^2 = 0.992$--$0.998$; appears flat at this scale). Physical slant, FOV, and curvature require progressively deeper processing, peaking around layers 17--19 before declining in the final layers. Embed = Conv3D patch projection output (before any transformer processing). Merger = post-transformer spatial merging ($2{\times}2$ tokens $\to$ 2048-dim). Ridge regression ($\alpha = 1.0$) on mean-pooled 1280-dim features (2048-dim for merger).}
    \label{fig:vlm_layerwise}
    \vspace{-0.2cm}
\end{figure}

\begin{figure}[t]
    \vspace{-0.25cm}
    \centering
    \includegraphics[width=1.0\linewidth]{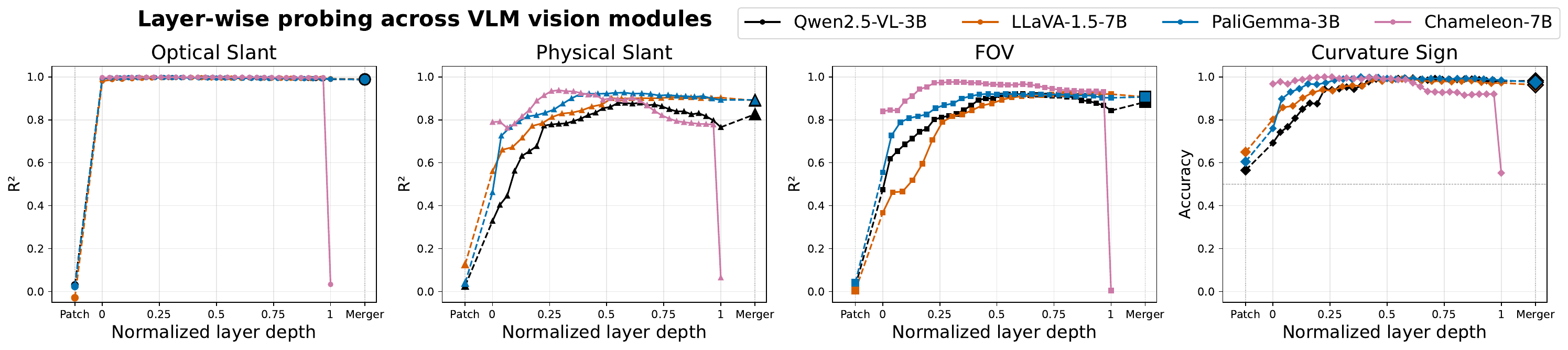}
    \vspace{-0.6cm}
    \caption{Layerwise probing across four VLMs vision modules, extending the single-model analysis in Qwen2.5-VL-3B to LLaVA-1.5-7B (CLIP-ViT-L/14), PaliGemma-3B (SigLIP-So400M), and Chameleon-7B (VQ-VAE). The result is consistent with the encoding being architecture-general, with the readout bottleneck arising at the language interface rather than in the vision tower.}
    \label{fig:layerwise_generalization}
    \vspace{-0.7cm}
\end{figure}

%% file: tables/vlm_vision_module_cmp.tex
\begin{table}[t]
\centering
\caption{Layer-wise probe for four VLMs with different vision encoders and language models. $R^2$ values listed for each geometric property at the layer where they peak.}
\label{tab:layerwise-probe-peaks}
\vspace{-0.4cm}
\scriptsize
\setlength{\tabcolsep}{1pt}
\resizebox{\linewidth}{!}{
\begin{tabular}{llllrrrr}
\toprule
Model & Vision tower & LM & Fusion & OS peak & FOV peak & PS peak & Curv peak \\
\midrule
Qwen2.5-VL-3B & Qwen-native ViT & Qwen2.5 & Late & 0.998 & 0.921 & 0.879 & 0.995 \\
LLaVA-1.5-7B & CLIP-ViT-L/14 & Vicuna & Late (frozen vision) & 0.997 & 0.923 & 0.905 & 0.990 \\
PaliGemma-3B & SigLIP-So400M & Gemma-2B & Late (prefix) & 0.997 & 0.921 & 0.926 & 1.000 \\
Chameleon-7B & VQ-VAE & unified & Early (vocab fusion) & 0.999 & 0.977 & 0.938 & 1.000 \\
\bottomrule
\end{tabular}
}
\vspace{-0.6cm}
\end{table}

%% file: figures/vlm_frozen_vision_regression.tex
% \begin{figure}[h]
%     \vspace{-0.4cm}
%     \centering
%     \includegraphics[width=0.32\linewidth]{images/rebuttal/prediction_distribution_qwen.png}
%     \includegraphics[width=0.32\linewidth]{images/rebuttal/prediction_distribution_llava.png}
%     \includegraphics[width=0.32\linewidth]{images/rebuttal/prediction_distribution_paligemma.png}
%     \vspace{-0.1cm}
%     \caption{Qwen, LLaVA, PaliGemma show continuous regression.}
%     \label{fig:token_regression}
%     \vspace{-0.25cm}
% \end{figure}

\begin{figure}[t]
    % \vspace{-0.4cm}
    \centering
    \includegraphics[width=0.72\linewidth]{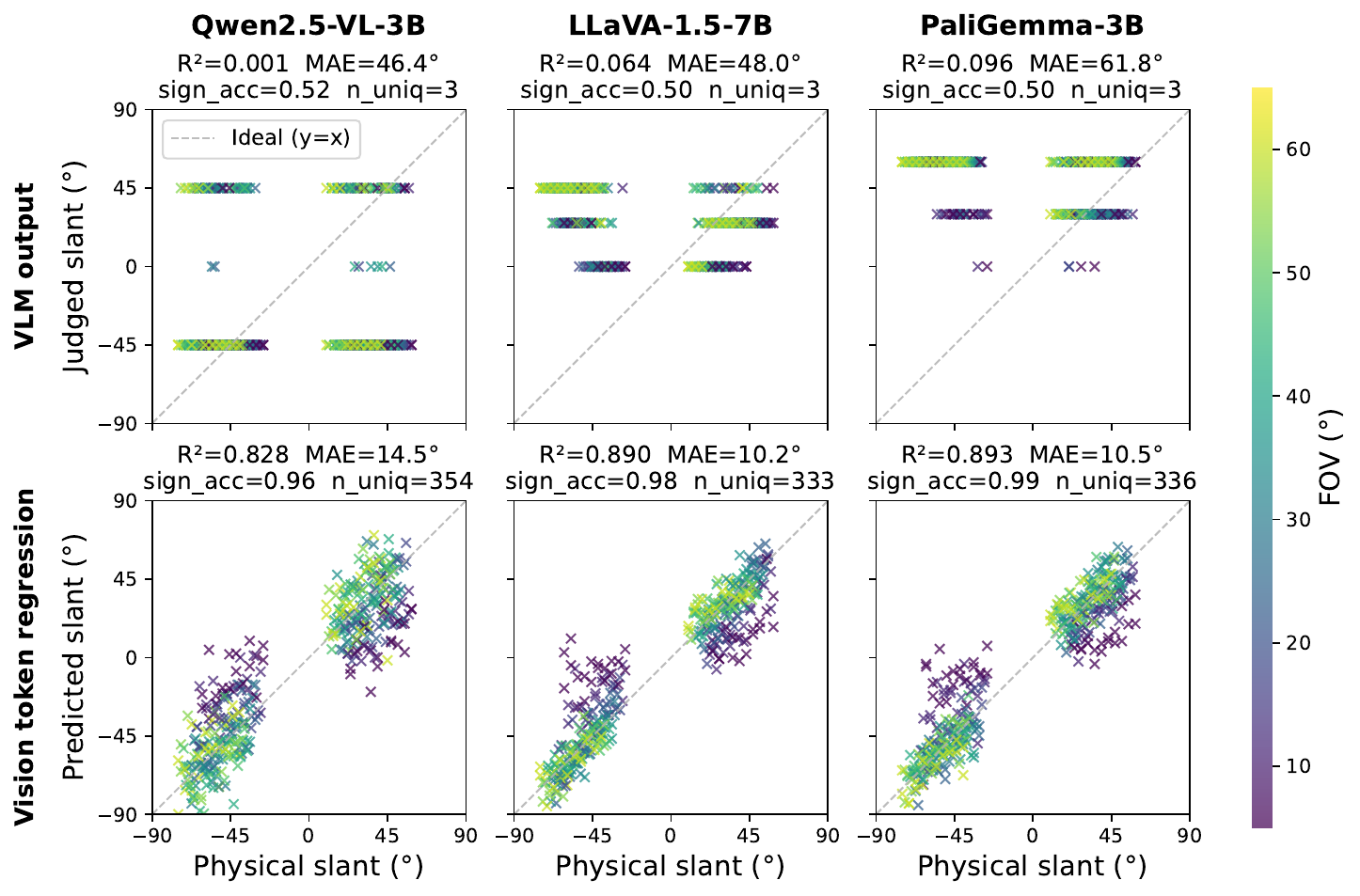}
    \vspace{-0.3cm}
    \caption{Readout bottleneck---VLM output anchors (top); LM-input vision tokens encode continuous slant (bottom). We extract the mean-pooled post-projector vision tokens (in LM embedding space) from Qwen, LLaVA, and PaliGemma and train a linear regressor on train image tokens to predict physical slant. Results on held-out test images.
    % Same model, same images, two readouts 
    }
    \label{fig:token_regression}
    \vspace{-0.3cm}
\end{figure}

%% file: figures/lm_ablation.tex
\begin{figure}[t]
    \centering

    % Qwen2.5-VL-7B
    % \includegraphics[height=2.5cm]{images/attention_ablation/qwen2.5-vl-3b_crop/baseline_scatter.png}
    % \includegraphics[height=2.5cm]{images/attention_ablation/qwen2.5-vl-3b_crop/sweep_scatter_L0_H01.png}
    % \includegraphics[height=2.5cm]{images/attention_ablation/qwen2.5-vl-3b_crop/sweep_scatter_L0_H00.png}
    % \includegraphics[height=2.5cm]{images/attention_ablation/qwen2.5-vl-3b_crop/sweep_scatter_L8_H26.png}

    % % Qwen2.5-VL-3B
    %  \includegraphics[height=2.5cm]{images/attention_ablation/qwen2.5-vl-3b_crop/baseline_scatter.png}
    % \includegraphics[height=2.5cm]{images/attention_ablation/qwen2.5-vl-3b_crop/sweep_scatter_L0_H01.png}
    % \includegraphics[height=2.5cm]{images/attention_ablation/qwen2.5-vl-3b_crop/sweep_scatter_L0_H00.png}
    % \includegraphics[height=2.5cm]{images/attention_ablation/qwen2.5-vl-3b_crop/sweep_scatter_L8_H04.png}
    % % \includegraphics[height=2.5cm]{images/attention_ablation/qwen2.5-vl-3b/colorbar.png}

    % % Qwen2-VL-7B
    % \includegraphics[height=2.5cm]{images/attention_ablation/qwen2-vl-7b_crop/baseline_scatter.png}
    % \includegraphics[height=2.5cm]{images/attention_ablation/qwen2-vl-7b_crop/sweep_scatter_L0_H09.png}
    % \includegraphics[height=2.5cm]{images/attention_ablation/qwen2-vl-7b_crop/sweep_scatter_L0_H23.png}
    % \includegraphics[height=2.5cm]{images/attention_ablation/qwen2-vl-7b_crop/sweep_scatter_L5_H19.png}
    % % \includegraphics[height=2.5cm]{images/attention_ablation/qwen2-vl-7b/colorbar.png}

    \includegraphics[width=0.85\linewidth]{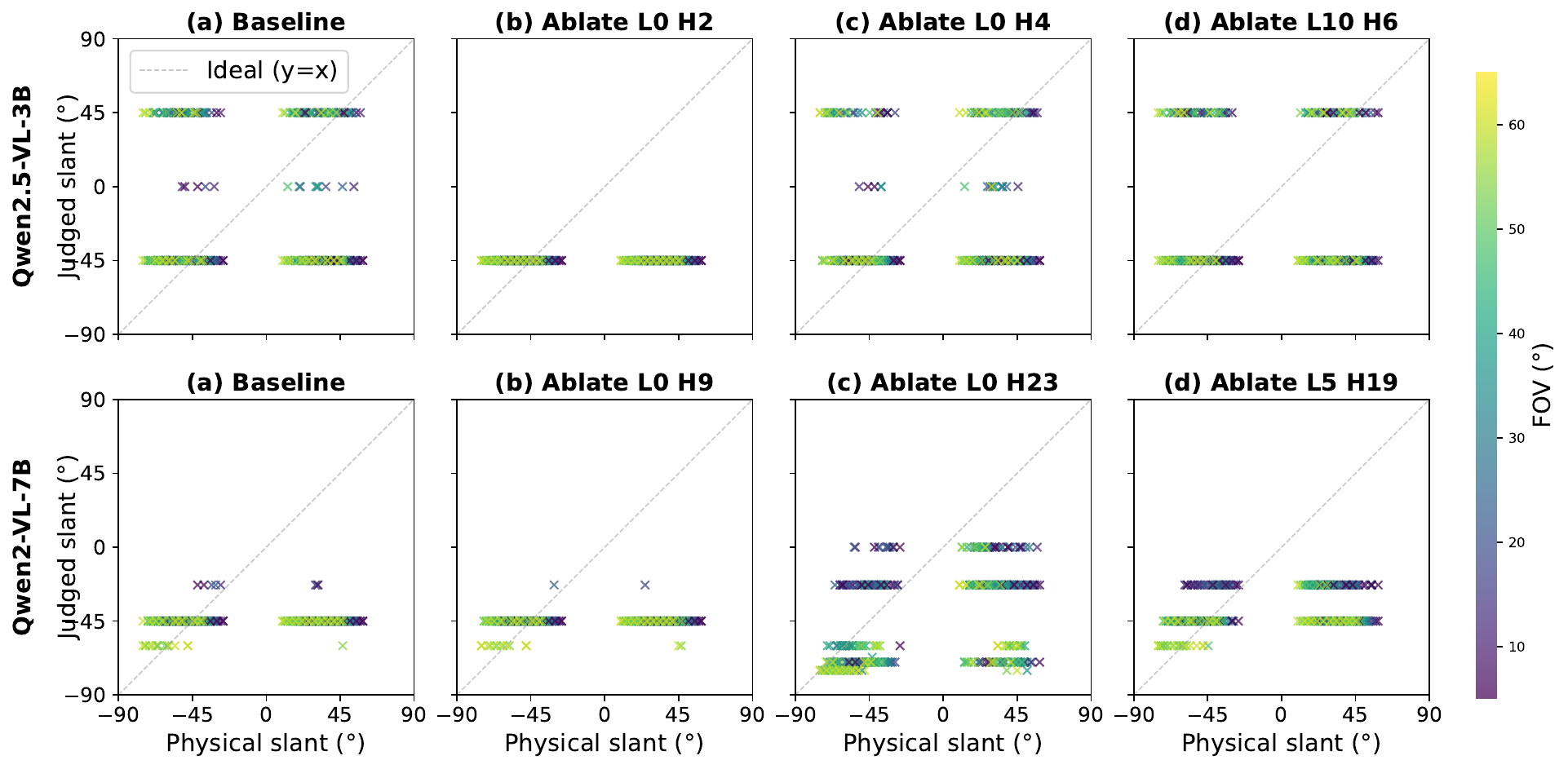}
    \vspace{-0.3cm}
    \caption{No single attention head ablation produces continuous stimulus-dependent predictions. We apply mean ablation on Qwen2.5-VL-3B (top) and Qwen2-VL-7B (bottom) slant prediction. 
    (a)~Baseline predictions; (b, c, d)~Typical results of ablating attention heads at different layers.
    The prediction barely changes; some even induced prompt copying, e.g.\ ablating L5\,H19, a new anchor at ``$-23^\degree$'' appears that matches the prompt example.
    % (a)~Baseline: predictions anchor at $-45^\degree$. (b)~Ablating L0\,H9: predictions barely change. (c)~Ablating L0\,H23: anchoring spreads across more discrete values. (d)~Ablating L5\,H19: a new anchor at $-23^\degree$ appears that matches the prompt example (prompt copying). 
    }
    \label{fig:lm_ablation}
    \vspace{-0.7cm}
\end{figure}

%% file: 4_results3_lm.tex
\subsection{Ablating Language Model Attention Heads}
\vspace{-0.15cm}

Does intervention reduce anchoring? Recent work by Rudman et al.~found individual attention heads in VLMs that copy prompts, and showed that ablating specific heads can eliminate copying of example values from the prompt~\cite{rudman2026mechanisms}. Inspired by this approach, we test whether anchoring in our task arises from specific attention heads in the language model. For each targeted head, we replace its pre-projection activations with the mean activation across all batch and token positions, removing position- and content-specific information while preserving activation scale.

We sweep all layer-head pairs in Qwen2.5-VL-3B (36 layers $\times$ 16 heads = 576 ablations), ablating one head at a time on the 400-stimuli test set.
% (per-head statistics in supplemental \cref{sec:ablation_details}).
No single head ablation eliminates anchoring (\cref{fig:lm_ablation}). Some ablations modestly reduce the dominance of the primary anchor value and increase response variance, but most simply redistribute responses across existing anchor values without producing graded estimates that vary continuously with stimulus parameters. %In a few cases, ablating a head increases prompt copying of the example value ($-23^\circ$), suggesting that some heads may suppress copying behavior.

We repeat the same analysis on the larger Qwen2-VL-7B model~\cite{qwen2vl_hf} (28 layers $\times$ 28 heads = 784 ablations) and observe a similar pattern: individual head ablations slightly redistribute responses but do not substantially reduce anchoring. These results suggest that anchoring in continuous perceptual estimation is not attributable to a single attention head but may instead reflect a distributed property of the language model. Although probing shows that geometric information is encoded in the vision representations, the language model cannot reliably read out continuous slant values. Simple attention-head interventions do not recover graded predictions.
If this bottleneck is distributed across heads, stronger interventions such as ablating head groups or steering activations along a learned slant direction remain for future work.

%% file: 5_conclusion.tex
\vspace{-0.25cm}
\section{Discussion}
\vspace{-0.15cm}

\paragraph{Anchoring as a language readout failure.}
Our central finding is not that VLMs perform poorly on slant-from-texture, but that they fail in a specific, structured way: predictions collapse to a small set of discrete values regardless of stimulus parameters, rather than scattering around the ground truth. This pattern is qualitatively distinct from humans and unsupervised CNNs, which map continuous stimulus variation to smooth, monotonic changes in perceived slant in a consistent, biased manner (curvature sign asymmetry B1, FOV dependency B2 and B3). The failure is therefore one of graded expression specifically, not of magnitude or precision. Our probing results (\cref{sec:results_vit}) localize this failure: geometric information is present and linearly decodable throughout the vision encoder and in the post-merger tokens passed to the language model, yet the language output does not reflect it. In effect, \emph{the model knows the geometry but does not say it.} Directly tracing how this information transforms through the language model's hidden states remains future work.

\vspace{-0.15cm}
\paragraph{Speculative: The cross-entropy bottleneck.}
A deeper question is whether the encoding-readout dissociation reflects a mismatch between training objectives.
VLMs are trained with next-token prediction (cross-entropy over discrete text tokens): a continuous value like $42.5^\circ$ must be expressed as a sequence of text tokens (``4'', ``2'', ``.'', ``5''), and the model must learn that this token sequence corresponds to a specific point on a continuous scale. 
There is no direct encouragement for the model to produce text outputs that co-vary smoothly with the continuous geometric variables encoded in vision tokens.
% Speculative: The anchoring behavior may therefore be an intrinsic limitation of the autoregressive text generation paradigm when applied to continuous perceptual variables, rather than a correctable readout failure.

\vspace{-0.15cm}
\paragraph{Speculative: Anchoring as a cue-deficient default value.}
Past human perception research found that perceived distance defaults toward a resting value when distance cues are not present~\cite{gogel1969sensing}. 
% (the specific distance tendency)
Our stimuli are deprived of any cues beyond texture gradients, and the models, trained on cue-rich images, may defer to a small set of ``default'' values. As shown in the probing results, texture-gradient information is present in the LM-input vision tokens, but the language module does not recover it, so the outputs fall back to the default values as they would under cue deficiency. SFT teaches the model to recover slant from these cue-deficient stimuli, reducing anchoring and improving the monotonicity of predictions.
Future work could test this by enriching the stimuli with additional geometric cues, to see whether graded predictions are recovered.

\vspace{-0.15cm}
\paragraph{Low FOV predictions.} Our stimuli are $224\times224$ pixels. Image resolution, dot size, and transformer tokenization block size likely affect the ability of the model to make predictions for low FOV stimuli because, at low FOV, texture gradients become small and so are ``hard to see.'' This relates to ideas of human visual acuity.

% \vspace{-0.15cm}
% \paragraph{Inference software variation.} Specific anchoring values were not consistent across Ollama inference backends (e.g., slight differences in numeric execution), though the amount of anchoring did not vary notably.

\vspace{-0.15cm}
\paragraph{Broader implications.}
\vspace{-0.10cm}
Our findings suggest a tension between language grounding and continuous geometric expression in the slant-from-texture task. VLMs perform well on high-level vision-language benchmarks that emphasize object recognition, scene understanding, and visual reasoning, yet they fail on a task that requires only the interpretation of low-level texture gradients. 
Standard VLM benchmarks emphasize categorical or semantic judgments that are naturally suited to discrete text output, and may therefore not reveal difficulties in expressing continuous variables. 
Slant-from-texture, grounded in psychophysical research, provides a simple but revealing test case for probing the boundaries of elementary visual understanding in multimodal models, especially when considering tasks in which VLMs must accurately predict what humans are likely to see.

%% file: 2_related.tex
\vspace{-0.25cm}
\section{Related Work}
\vspace{-0.15cm}
\label{sec:related}

\paragraph{Human perception of slant-from-texture.}
Slant from texture is a classic perceptual problem with well documented human biases.
Oru{\c{c}} et al.~\cite{orucc2003weighted} compared linear perspective (a grid of lines) and texture gradient (diamond-shaped texture elements) cues at 75\degree~slant. Observers performed better with combined cues than with either cue alone. The results were consistent with a linear combination of estimates from cues.
Todd et al.\ showed that human slant-from-texture perception is systematic and biased~\cite{todd2005effects,todd2007effects}, identifying consistent effects of field of view and curvature sign (biases B1-B3 in our notation).
Recent work further isolates which texture cues drive slant, comparing cue-conflict and cue-consistent conditions.
Chen et al.~found that texture compression influenced slant settings more than other texture cues, with its influence decreasing with larger field of view (10\degree~vs.\ 20\degree) and less regular textures~\cite{chen2020multiple}.
Among texture types, regular blob (polka-dot) patterns are popular for modeling texture perception~\cite{knill1998surface} and produce superior slant discrimination performance in humans compared to uniform lattices, Voronoi tessellations, plaids, and noise~\cite{rosas2004some}. Unlike plaids or contour textures, they lack explicit perspective lines, making them a purer test of texture-gradient processing. They can also be systematically manipulated to test the effects of gradient compression~\cite{chen2020multiple} and gradient disruption~\cite{kemp2024sensory}.

\vspace{-0.15cm}
\paragraph{Computational models of slant-from-texture perception.} 
CNNs trained on texture statistics reproduce human-like slant biases under unsupervised learning objectives~\cite{wang2023human}, supporting the use of deep networks as computational analogues for cue learning and inference.
This connects to classic shape-from-texture formulations that recover surface orientation from texture gradients under assumptions such as homogeneity~\cite{witkin1981recovering,kanatani1989shape}, and to recent computer vision models that revisit this problem under realistic settings (unknown textures, nuisance factors), showing texture alone can constrain local surface geometry in the wild~\cite{verbin2020toward}.
In parallel, modern monocular depth estimation learns strong geometric priors from large-scale data, both supervised and self-supervised~\cite{eigen2014depth,godard2019digging,yang2024depthanything}, but these systems typically entangle multiple cues and do not isolate the specific contribution of texture gradients to surface slant.

\vspace{-0.15cm}
\paragraph{Vision-language models for visual perception.}
VLMs couple a pretrained vision encoder with a large language model via a lightweight alignment module, enabling strong performance on broad multimodal understanding and reasoning benchmarks~\cite{liu2024mmbench, yue2024mmmu, lu2024mathvista, li2024llava, bai2023qwen, kamath2025gemma}. However, their behavior on low-level perceptual and geometric tasks (e.g., fine-grained spatial relations, orientations, and viewpoint-dependent reasoning) is less well understood; existing spatially focused evaluations reveal substantial gaps~\cite{liu2023vsr,yu2025sibench}.
Emerging diagnostics suggest that large vision/VLM models exhibit weaknesses in metric depth and geometry understanding too, motivating targeted probes and RGB-D augmentation~\cite{danier2024depthcues,cai2025spatialbot,cheng2024spatialrgpt}.
Overall, many training and evaluation approaches emphasize semantic recognition and language generation (captioning/VQA/grounding) rather than geometry.
% \MG{cite works}

\vspace{-0.15cm}
\paragraph{Failure modes of VLMs.}
Controlled perceptual evaluations of VLMs remain rare. Zhang et al.~\cite{zhang2023grounding} test five visual illusions (color constancy, color assimilation, color contrast, geometrical relativity, and geometrical perspective), but most studies use naturalistic images.
More broadly, a growing body of work documents systematic VLM failures, particularly on tasks requiring precise visual grounding and geometric reasoning. VLM failures on geometric primitives have been shown to depend upon spurious correlations rather than image evidence~\cite{rudman2025forgotten,rahmanzadehgervi2024vision, tong2024eyes,vo2025vision, huang2025vision, zhang2024mathverse, gao2023g}. Mechanistic analyses suggest that some errors stem from over-reliance on language signals, including anchoring to prompt-provided cues and copying textual constraints even when they conflict with visual evidence~\cite{rudman2026mechanisms}. However, much of this literature relies on naturalistic images, leaving open the question of how VLMs behave under tightly controlled psychophysical-style stimuli that isolate low-level geometric cues. 
\CC{
Zhang et al.~\cite{zhang2024visually} study image classification failures of VLMs. Their work provides tangential evidence for ``underutilized visual features'' in VLMs, considering classification where the output space is already discrete. We study a continuous geometric quantity, and find anchoring to a small set of salient values independent of stimulus parameters, which has no analog in discrete classification. 
}

%The findings from prior works motivate controlled evaluations that directly probe whether VLMs encode and use texture-based slant information.

%\MG{I know this is work in progress but leaving the comment here to remember. I think since the framing of the paper and title is about VLM failures, you should cite other works that talk about VLMs failing, especially in geomteric contexts. This is how i would go about it (just an example skeleton): 
%Prior works have shown VLMs fail at linguistic tasks (cite "NegVQA: Can Vision Language Models Understand Negation?"), but a much larger body of work has shown visual failures (cite Forgotten Polygons, VLMs are Blind, VLMs are Biased, Eyes Wide Shut, Why Vision Language Models Struggle with Visual Arithmetic?, and even the 4 geometric papers i sent). Most have attributed the failures to "blindness" of the vision encoder and some attribute it to over reliance on pre-training data. However, much of the work out there relies on linguistic cues from the prompt or in the image, etc. Something along those lines. }

%% file: 7_supplemental.tex
% \clearpage
% \appendix
\setcounter{page}{1}
% \maketitlesupplementary
\setcounter{section}{0}

% \section{Contextual comparison with prior work}
% We include this figure to provide context for how the present experiments relate to prior human and CNN studies~\cite{todd2005effects,wang2023human}; it is not used to support any quantitative claims in the main paper.
% \input{figures/comparative_analysis.tex}
% % \input{tables/comparative_analysis.tex}

% \section{TODOs}
% \subsection{Code}
% \begin{enumerate}
%     \item Figure 1~\ref{fig:results_technical}: results of some models missing
%     \item Figure 2~\ref{fig:results_incontext}: re-plot
%     \item normal train/test split for SFT model results (currently reversed in main text to show the more interesting results, but may be confusing to readers)
% \end{enumerate}

% \subsection{Writing}
% \begin{enumerate}
%     \item What's the best way to organize those sections in supp?
%     \item write about ollama and hugging face models. Include a comparison using qwen2.5-vl-3b?
%     \item boxplots of each model per prompt type? Maybe not, deprioritizing this.
% \end{enumerate}

%-------------------------------------------------------------------------------
\section{Recent models}
% Added from rebuttal:
\input{figures/results_natural_frontierVLMs_scatterplot.tex}

\CC{
We also found anchoring in 12 frontier VLMs, including closed-source ones. We evaluated InternVL3.5-1B~\cite{wang2025internvl3}, PaliGemma-3B~\cite{beyer2024paligemma}, LLaVA-1.5-7B~\cite{llava_hf}; Gemini-3.1-Flash-Lite~\cite{Gemini3.1FlashLite}, Qwen3.5-4B~\cite{qwen3.5}, Qwen3.6-35B-A3B (MoE)~\cite{qwen36_35b_a3b}; Claude-Haiku-4.5, Sonnet-4.6, Opus-4.7~\cite{Anthropic2026Claude}; GPT-4o~\cite{hurst2024gpt}, 5.4-nano, 5.4~\cite{gpt54_openai} on the 400-stimulus test set with our natural-language prompt. \emph{Every model} exhibits anchoring: their predictions collapse onto a few unique values vs.~200 distinct ground-truth physical-slant levels (\cref{fig:frontier_scatter}).
}

\CC{
For example, Qwen3.6-35B-A3B concentrates 274/400 predictions on exactly $0^\circ$; Gemini-3.1-Flash-Lite places 301/400 at exactly $-45^\circ$; GPT-4o has 263/400 on just 0\degree~and -45\degree. $R^2$ with physical slant is $\leq 0.16$ for every model (mean 0.04), and sign-of-curvature accuracy is at chance (0.46--0.58). Thus, the anchoring bottleneck is not specific to open-weight or older VLMs, nor to a particular family (Gemini, Qwen, LLaVA, PaliGemma); all show the issue.
GPT models show a slightly different pattern with more variability and less extreme anchoring, suggesting some improvement. But slant estimation MAE remains high, and curvature sign discrimination is near chance.
}

%-------------------------------------------------------------------------------
\section{Prompt design and output formats}
We include full prompt specifications to support reproducibility and to demonstrate that anchoring persists across substantial linguistic variation.

Full prompt text for each style of regression task: natural~\cref{tab:prompt_natural}, technical~\cref{tab:prompt_technical}, and in-context modifiers~\cref{tab:prompt_incontext}. SFT uses natural language prompts. Binary classification task also adopts natural language style prompts for consistency~\cref{tab:prompt_binary}.

Note: earlier experiments included confidence ratings and free-text reasoning to assess response variability under stochastic decoding. As results showed that predicted slant values were strongly anchored and largely insensitive to decoding temperature, subsequent analyses focus on slant estimates alone.

%-------------------------------------------------------------------------------
\section{Additional plots and anchoring analyses}
In addition to natural language prompts (regression and binary classification), we include heatmaps for other prompt types (~\cref{fig:results_technical} and ~\cref{fig:results_incontext})

\input{figures/results_technical.tex}
\input{figures/results_incontext.tex}
% \input{figures/results_binary.tex}

% Temperature robustness (results not included yet)
% % \input{tables/results_temperature.tex}

%-------------------------------------------------------------------------------
% \section{SFT train/test sets}
% In main text, we reverse the train and test sets, reporting results for SFT models trained on 400 test images and evaluated on the 2000 train images. 

% Here we show results on the original train/test split, where SFT models are trained on 2000 images and evaluated on 400 test images. We also generated additional test sets with 8 different random seeds (200x8=1600 images) and evaluate on them for clarification in case of confusion about the unusual reversing decision (results not included in main text). We find that the asymmetric behaviors for concave and convex images persist.

%-------------------------------------------------------------------------------
% \paragraph{Synthetic polka-dot OOD concern.}
% \CC{Our polka-dot stimuli are intentionally controlled to isolate texture-gradient cues. While they may or may not be OOD (we do not always know the VLM training set), the OOD framing does not by itself explain the qualitative shape of failures: SFT models still show anchoring effects, and sign-of-curvature accuracy is near chance even when sign is the only required output. This pattern indicates readout discretization, not OOD generalization failure; this is supported by geometric probing.}

\section{Are failures due to out-of-distribution stimuli?}

Our stimuli are synthetic polka-dot textures rendered under controlled conditions to isolate texture-gradient cues. A natural concern is that these images are out of distribution (OOD) relative to VLM training data, which consists predominantly of natural photographs and Web content. They may or may not be OOD for the VLMs we test, as we do not often know the VLM training set. But, if we suppose they are, then one might argue that the observed anchoring failures reflect a simple OOD generalization failure rather than a principled readout bottleneck at the language interface.

Two observations argue against a pure OOD account. 
First, linear probing confirms that the vision encoder extracts slant-relevant information from these stimuli at high fidelity, ruling out a failure of low-level visual processing. Polka-dot and regular blob textures are likely to appear in natural scenes and manufactured objects (though not in their cue-deficient form) within large-scale Web-crawled training data.
Second, the anchoring failure pattern persists after supervised fine-tuning (SFT) on the task stimuli: even after training on thousands of polka-dot images, anchoring is not eliminated and sign-of-curvature errors remain. An OOD explanation would predict that SFT, by familiarizing the model with the stimulus distribution, should largely remedy the failures. We note that our SFT uses LoRA on the q/v projections only (main paper, footnote 1); however, the linear regression-head result shows the post-projector features are already slant-decodable, so the failure to remedy anchoring is not attributable to insufficient fine-tuning capacity.
%
%Third, sign-of-curvature accuracy is near chance even when curvature sign is the only required output, a binary discrimination that generic OOD degradation would not selectively impair. 

Taken together, these observations are consistent with the failures reflecting a readout bottleneck in the language model rather than an inability of the vision encoder to process the stimuli.

At a higher level, if a VLM is to be a `foundation model,' especially for interactions with humans, then we should want it to generalize to basic stimuli \emph{even if they are OOD}. Polka dot patterns like ours are rare in nature and arguably OOD: they have no shading cues, only texture gradient cues. And yet human vision is easily capable of estimating shape from polka dot texture cues. So, our setup is meaningful when considering the broader goals of building AI systems that can understand and predict human perception when necessary.

%-------------------------------------------------------------------------------
\section{Inference backend variation: Ollama vs.\ HuggingFace}
We ran a subset of experiments using two different inference backends for the same underlying model (Qwen2.5-VL-3B): Ollama and Hugging Face.
The two backends produce numerically different anchor values---for example, Ollama may produce a dominant anchor at $-45^\circ$ while Hugging Face produces anchoring at a different discrete value---likely due to differences in quantization, sampling defaults, or tokenizer handling between backends.
Critically, however, the anchoring \emph{pattern} is consistent: in both cases, predictions collapse to a small set of discrete values and do not co-vary systematically with stimulus parameters.
All primary results reported in the main paper use Hugging Face Qwen models, which provides direct access to model weights without additional quantization.

% TODO: add side-by-side heatmap comparing Ollama vs. HF for Qwen2.5-VL-3B (natural prompt, same 400 images)

%-------------------------------------------------------------------------------
\section{Attention head ablation details}
\label{sec:ablation_details}

% TODO: fill in per-model details

\paragraph{Qwen2.5-VL-3B} (36 layers $\times$ 14 heads = 504 ablations).
Baseline predictions collapse to two discrete anchor values with no prompt copying.
Ablating individual heads merely redistributes which images anchor to which value without spreading the response distribution or changing the anchor values.
% TODO: add full sweep heatmap

\paragraph{Qwen2-VL-7B} (28 layers $\times$ 28 heads = 784 ablations).
Baseline predictions anchor at $-45^\circ$ with mode percentage 94\% and correlation $R^2=0.022$ ($r = 0.149$) with ground-truth physical slant.
\input{tables/prompt_natural.tex}
\input{tables/prompt_technical.tex}

\input{tables/prompt_binary.tex}
% \input{tables/prompt_naturalnoreasoningincontext.tex}
% \input{tables/prompt_incontext.tex} % not used since incorporated into prompt_binary.tex

% \paragraph{Distributions of error and mode per model.} Boxplots in \cref{fig:results_natural} show the per model distributions of slant estimates for natural language prompts. Heatmaps in \cref{fig:results_technical} and \cref{fig:results_binary} provide a visual summary of the anchoring effects with prompt variations.

% % \paragraph{Result tables.} \cref{tab:results_natural}, \cref{tab:results_technical}, \cref{tab:results_binary}, \cref{tab:results_naturalincontext} show the results for slant estimation with different prompt families. 

% % \input{figures/sft_training.tex}

% The severe anchoring effects prevent VLMs from showing any relationship between accuracy and field of view or optical slant, as in \cref{fig:acc_by_optical_1}, \cref{fig:acc_by_optical_2}, \cref{fig:acc_by_optical_3}, \cref{fig:acc_by_optical_4}.
% \input{figures/acc_by_optical_0825-120542_natural.tex}

% % \input{tables/results_naturalnoreasoningincontext.tex}
% % \input{tables/results_naturalincontext.tex}
% % \input{tables/results_error.tex}
% % \input{tables/df_pred_gt.tex}
% % \input{tables/datasets.tex}
% % \input{tables/experimental_factors.tex}

%% file: figures/results_natural_frontierVLMs_scatterplot.tex
\begin{figure}[b]
    % \vspace{-0.65cm}
    \centering
    \includegraphics[width=1.0\linewidth]{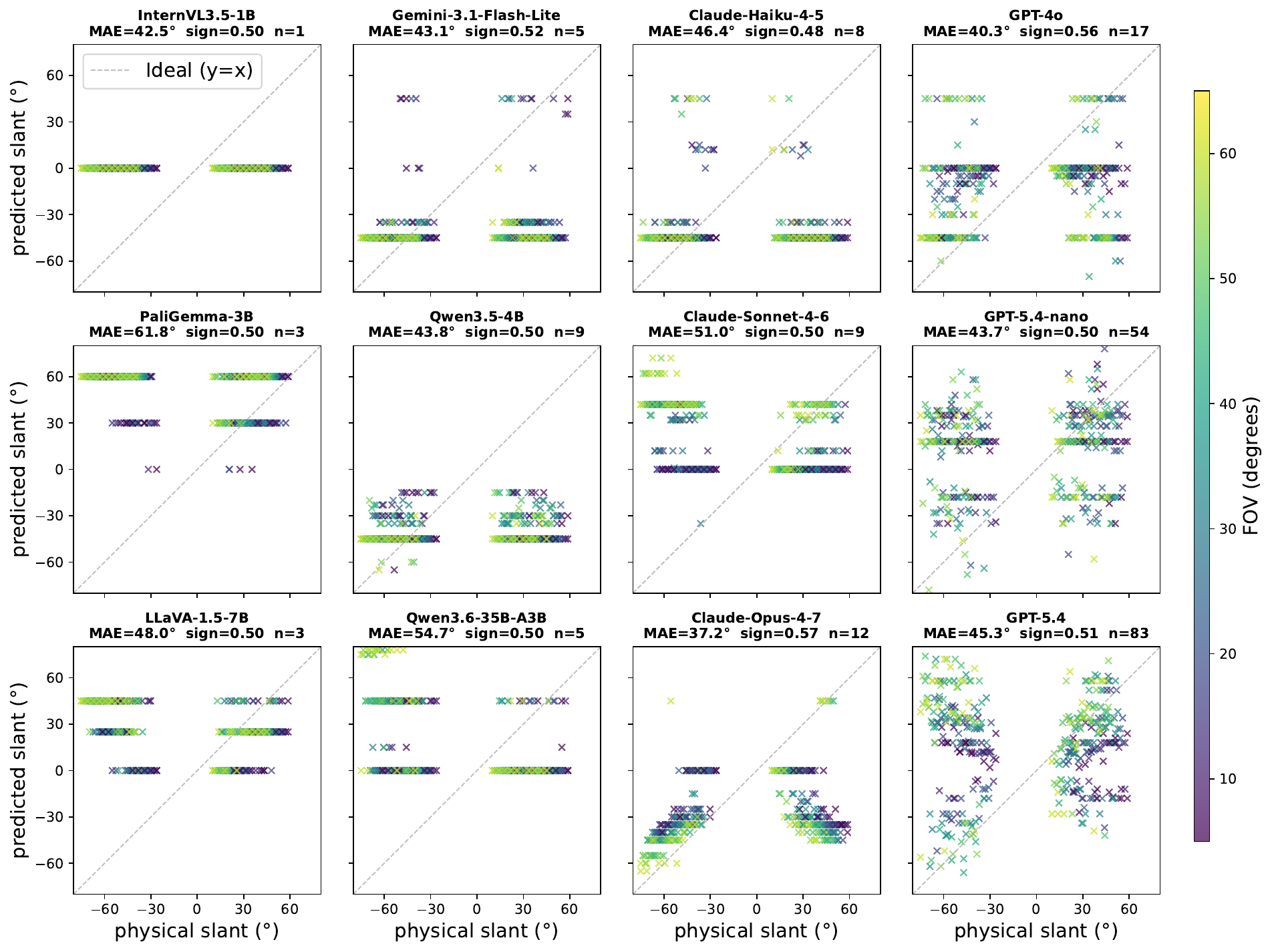}
    \vspace{-0.5cm}
    \caption{Scatter plot of model predictions across 12 recent VLMs. Each panel shows predictions from a different model, with mean absolute error (MAE) for physical slant, accuracy for curvature sign discrimination, and the number of unique predicted values (out of 400 stimuli) in the title.}
    \label{fig:frontier_scatter}
    \vspace{-0.25cm}
\end{figure}

%% file: figures/results_technical.tex
\begin{figure*}[h]
\begin{center}
   \includegraphics[width=0.9\linewidth]{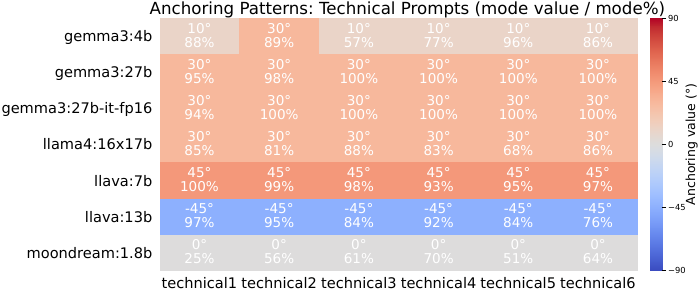}
\end{center}
\caption{Under technical prompt settings, mode value and frequency of results across models in heatmap.}\label{fig:results_technical}
\end{figure*}

%% file: figures/results_incontext.tex
% \begin{figure}[t]
% \begin{center}
%    \includegraphics[width=1.0\linewidth]{images/incontext/json.png}
% \end{center}
% \caption{Comparison of slant prediction error of VLM estimations from in context learning. }\label{fig:results_incontext_json}
% \end{figure}

% \begin{figure*}[b]
% \begin{center}
%    \includegraphics[width=1.0\linewidth]{images/incontext/prompt_detail.png}
% \end{center}
% \caption{Comparison of slant error metrics by prompt asking. }\label{fig:results_incontext_prompt_detail}
% \end{figure*}

% \begin{figure*}[h]
% \begin{center}
% %    \includegraphics[width=1.0\linewidth]{images/natural/boxplot_natural.png}
% %    \includegraphics[width=1.0\linewidth]{images/natural/boxplot_json.png}
%    \includegraphics[width=0.5\linewidth]{images/incontext/heatmap.png}
% \end{center}
% \caption{For all models and prompt settings, averaged performance of models in boxplots and mode value and frequency in heatmap.}\label{fig:results_incontext}
% \end{figure*}

\begin{figure*}[h]
\begin{center}
   \includegraphics[width=0.8\linewidth]{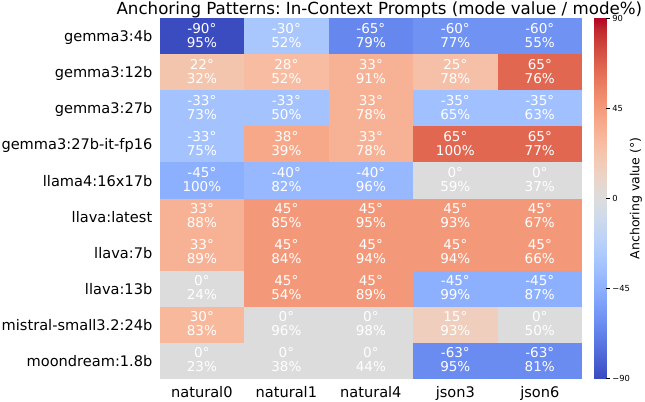}
\end{center}
\caption{With the in-context modifiers and natural language prompt settings, mode value and frequency of results across models in heatmap.}\label{fig:results_incontext}
\end{figure*}

%% file: tables/prompt_natural.tex
\begin{table*}[t]
\caption{VLM prompts in the ``natural language'' style, including variants and their components. Style changes in blue.
% See \cref{tab:results_natural} for results.
}
\label{tab:prompt_natural}
\begin{subtable}{\textwidth}
\centering
\begin{tabular}{p{0.2\linewidth} p{0.8\linewidth}}
\toprule
Components & Prompt \\
\midrule
setup & The image shows two connected flat surfaces forming \textcolor{codegreen}{a folded shape, like a book or greeting card.} Both surfaces are covered with identical round polka dots. The fold creates either a \textcolor{codegreen}{`valley' (concave, opening away from you) or a `ridge' (convex, pointing toward you).} \\
\midrule
task& Analyze the polka dot distortions to estimate the \textcolor{codegreen}{surface slant angle.} IMPORTANT: Your angle estimate MUST be \textcolor{codegreen}{between -90 and +90 degrees.} -90° = deepest possible valley (surfaces nearly horizontal pointing away), 0° = flat vertical surface, +90° = sharpest possible ridge (surfaces nearly horizontal pointing toward you). Also estimate your confidence between 0.0 and 1.0. \\
\midrule
cues&\textcolor{codegreen}{Visual guide:} Valley shapes compress and shrink dots near the fold center. Ridge shapes stretch and enlarge dots near the fold center. Stronger angles create more dramatic dot distortions. \\
\midrule
format&Response format: (angle, confidence); brief\_reasoning. Constraints: angle between -90 and +90, confidence between 0.0 and 1.0. Begin your response immediately with the opening parenthesis. \\
\midrule
format\_eg&Response format: (angle, confidence); brief\_reasoning \\
&\textcolor{codegreen}{Example:} (-23, 0.8); Dots are compressed near center indicating valley shape. Constraints: angle between -90 and +90, confidence between 0.0 and 1.0. Start your response immediately with the parentheses. No other text before it. \\
\midrule
format\_eg\_json&Respond in this exact \textcolor{codegreen}{JSON format}: \{"angle": number, "confidence": number, "reasoning": "text"\}. Constraints: angle between -90 and +90, confidence between 0.0 and 1.0. \textcolor{codegreen}{Example:} \{"angle": -23, "confidence": 0.8, "reasoning": "Dots compressed near fold"\}\\
\midrule
prompt\_minimal no\_anchor&Analyze this polka-dot folded surface. Valleys compress dots near the fold, ridges stretch them. Estimate slant angle (MUST be -90 to +90 degrees) and confidence (0.0-1.0). Response format: (angle, confidence); brief\_reasoning. Begin immediately with the opening parenthesis. \\
\bottomrule
\end{tabular}
\caption{Prompt components and their details}
\label{tab:prompt_components_natural}
\end{subtable}

\vspace{1em}

\begin{subtable}{\textwidth}
\centering
\scriptsize
% \begin{tabular}{p{0.06\linewidth} p{0.6\linewidth}}
\begin{tabular}{l l}
\toprule
Prompt & Components Included \\
\midrule
0 & prompt\_minimal \\
1 & setup + task + format\ \\
2 & setup + task + format\_eg \\
3 & setup + task + format\_eg\_json \\
4 & setup + task + cues + format\ \\
5 & setup + task + cues + format\_eg \\
6 & setup + task + cues + format\_eg\_json \\
\bottomrule
\end{tabular}
\caption{Prompt variants and their component combinations.}
\label{tab:prompt_variants_natural}
\end{subtable}
\end{table*}

%% file: tables/prompt_technical.tex
\begin{table*}[t]
\caption{VLM prompts in the ``technical language'' style, with variants and their components. Style changes in blue.
% See \cref{tab:results_technical} for results.
}
\label{tab:prompt_technical}
\begin{subtable}{\textwidth}
\centering
\begin{tabular}{l p{0.8\linewidth}}
\toprule
Components & Prompt \\
\midrule
setup & The image shows a scene with two slanted, \textcolor{codegreen}{geometrically symmetric}, flat surfaces textured with round polka dots of the same size. The slant is defined as the \textcolor{codegreen}{dihedral angle} of the surfaces relative to the \textcolor{codegreen}{fronto-parallel plane}. A \textcolor{codegreen}{convex} shape points toward the viewer, while a \textcolor{codegreen}{concave} shape points away.\\
\midrule
task\_signedslant & Please estimate the slant of the surface in degrees, within the range [-90, 90], where negative values indicate concave and positive values indicate convex. \\
\midrule
format&IMPORTANT: Your answer MUST follow the format (slant\_degree). Constraints: angle\_number between -90 and +90. Begin your response immediately with the opening parenthesis. Do not include any extra text before or after your answer.\\
\midrule
cues&Additional information: The slant will cause distortions in the polka dots, depending on both the sign and the magnitude of the slant angle. The greater the slant angle, the more distorted the dots will appear.\\
\midrule
variations&The input image is from a set that includes both convex and concave shapes, with various surface slant angles and different fields of view. \\
\midrule
effects&Near the intersection of the two planes, for a fixed field of view: the larger the convex angle, the larger and more stretched the dots appear. The larger the concave angle, the smaller and more compressed the dots appear. For a fixed slant angle, increasing the field of view increases the distortion of the dots. \\
\midrule
biases&\textcolor{codegreen}{Previous psychophysical experiments} have shown that humans consistently underestimate the slant of surfaces, especially for concave shapes. The accuracy of estimation is also affected by the field of view: smaller fields of view result in a larger underestimation bias. \\
\midrule
hint&\textcolor{codegreen}{Recent research} shows that when trained on the same stimulus images, unsupervised CNNs exhibit a similar bias to humans at test time, regardless of the neural network architecture. You may use this knowledge to adjust your estimation.\\
\bottomrule
\end{tabular}
\caption{Prompt components and their details.}
\label{tab:prompt_components_technical}
\end{subtable}

\vspace{1em}

\begin{subtable}{\textwidth}
\centering
\scriptsize
% \begin{tabular}{p{0.06\linewidth} p{0.93\linewidth}}
\begin{tabular}{l l}
\toprule
Prompt & Components Included \\
\midrule
1 & setup + task\_signedslant + format\_signedslant \\
2 & setup + task\_signedslant + format\_signedslant + cues \\
3 & setup + task\_signedslant + format\_signedslant + cues + variations \\
4 & setup + task\_signedslant + format\_signedslant + cues + variations + effects \\
5 & setup + task\_signedslant + format\_signedslant + cues + variations + effects + biases \\
6 & setup + task\_signedslant + format\_signedslant + cues + variations + effects + biases + hint \\
\bottomrule
\end{tabular}
\caption{Prompt variants and their component combinations.}
\label{tab:prompt_variants_technical}
\end{subtable}
\end{table*}

%% file: tables/prompt_binary.tex
\begin{table}[t]
\caption{VLM prompt modifiers for in context settings. Style changes in blue.
}
\label{tab:prompt_incontext}
\begin{subtable}{\textwidth}
\centering
\begin{tabular}{p{0.15\textwidth} p{0.15\textwidth} p{0.15\textwidth} p{0.15\textwidth} p{0.15\textwidth} p{0.15\textwidth}}
\toprule
\textcolor{codegreen}{In-context learning prompt} & \multicolumn{5}{>{\fontsize{8}{10}\selectfont\ttfamily}p{0.8\linewidth}}{You are about to see a few example images and one test image (the last one). The true labels for the examples are as follows: (-62.778\degree, 1.0); (-49.444\degree, 1.0); (49.583\degree, 1.0); (37.222\degree, 1.0).} \\
& \multicolumn{5}{>{\fontsize{8}{10}\selectfont\ttfamily}p{0.8\linewidth}}{Now here is the test image. This image shows ...} \\
\midrule
\textcolor{codegreen}{Images} & Example 1 & Example 2 & Example 3 & Example 4 & Test Image \\
% \midrule
&
\includegraphics[width=\linewidth]{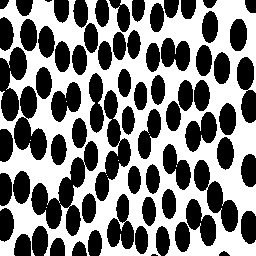} &
\includegraphics[width=\linewidth]{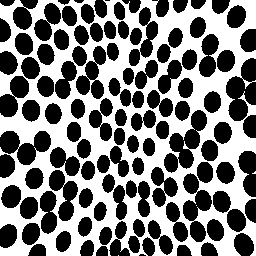} &
\includegraphics[width=\linewidth]{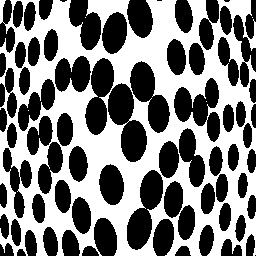} &
\includegraphics[width=\linewidth]{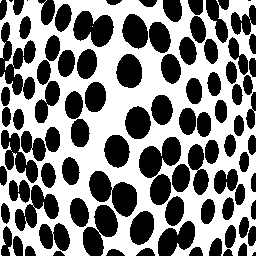} &
\includegraphics[width=\linewidth]{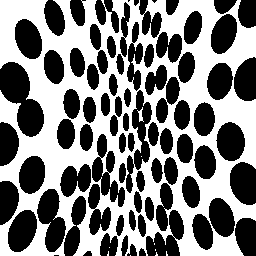} \\
\midrule
\textcolor{codegreen}{GT labels} &
\fontsize{8}{10}\texttt{Concave, Physical slant -62.778\degree, FOV 11.11\degree} &
\fontsize{8}{10}\texttt{Concave, Physical slant -49.444\degree, FOV 35.56\degree} &
\fontsize{8}{10}\texttt{Convex, Physical slant 49.583\degree, FOV 41.67\degree} &
\fontsize{8}{10}\texttt{Convex, Physical slant 37.222\degree, FOV 60.00\degree} &
\fontsize{8}{10}\texttt{Concave, Physical slant 56.111\degree, FOV 53.89\degree} \\
\bottomrule
\end{tabular}
\caption{Example in-context learning setup with labeled images and labels for few-shot slant estimation. The example images, test image, and prompts are sent to the model together in one query.}\label{tab:prompt_incontext_stimulilabels}
\end{subtable}

\vspace{1em}

\begin{subtable}[t]{\textwidth}
\centering
\scriptsize
% \begin{tabular}{p{0.06\linewidth} p{0.4\linewidth}}
\begin{tabular}{l l}
\toprule
Prompt & Components Included \\
\midrule
0 & prompt\_minimal\\
1 & setup + task + format \\
3 & setup + task + format\_json \\
4 & setup + task + cues + format \\
6 & setup + task + cues + format\_json \\
\bottomrule
\end{tabular}
\caption{Prompt variants and their component combinations.}\label{tab:prompt_variants_incontext_no_reasoning}
\end{subtable}

\end{table}

\begin{table}[t]
\caption{VLM prompts for binary settings, with their variant components. Style changes in blue.
}
\label{tab:prompt_binary}

\begin{subtable}{\textwidth}
\centering
\begin{tabular}{p{0.2\linewidth} p{0.8\linewidth}}
\toprule
Components & Prompt \\
\midrule
setup&Now here is the test image. This image shows two connected flat surfaces with polka dots, forming a folded shape. The fold creates either a ``valley'' (like an open book) or a ``ridge'' (like a roof peak). \\
\midrule
task&\textcolor{codegreen}{Determine whether this shows a VALLEY or RIDGE.} \\
\midrule
visual\_cues&Key visual indicators: In VALLEY configurations, polka dots near the fold appear smaller and more compressed. In RIDGE configurations, polka dots near the fold appear larger and more stretched. \\
\midrule
format&Response format: (VALLEY/RIDGE). Begin your response immediately with the opening parenthesis. \\
\midrule
cues&Visual guide: Valley shapes compress and shrink dots near the fold center. Ridge shapes stretch and enlarge dots near the fold center. \\
\midrule
prompt\_minimal&Look at this folded surface with polka dots. Is it folded like a VALLEY (inward fold, like an open book) or a RIDGE (outward fold, like a roof)? Response format: (\textcolor{codegreen}{VALLEY/RIDGE}). \\
\bottomrule
\end{tabular}
\caption{Prompt components and their details}\label{tab:prompt_binary_prompts}
\end{subtable}

\vspace{1em}

\begin{subtable}[t]{\textwidth}
\centering
\scriptsize
% \begin{tabular}{p{0.06\linewidth} p{0.4\linewidth}}
\begin{tabular}{l l}
\toprule
Prompt & Components Included \\
\midrule
0 & prompt\_minimal \\
1 & setup + task \\
2 & setup + task + visual\_cues \\
\bottomrule
\end{tabular}
\caption{Prompt variants and their component combinations}\label{tab:prompt_binary_variants}
\end{subtable}
\end{table}